\documentclass[10pt,twocolumn,letterpaper]{article}

\usepackage{cvpr}              
\usepackage{xcolor}
\usepackage[table]{xcolor}
\usepackage{colortbl, booktabs, pifont, multirow, gensymb}
\definecolor{cvprblue}{rgb}{0.21,0.49,0.74}
\usepackage[pagebackref,breaklinks,colorlinks,allcolors=cvprblue]{hyperref}

\def\paperID{*****} 
\def\confName{CVPR}
\def\confYear{2026}

\title{Fourier Angle Alignment for Oriented Object Detection in Remote Sensing}

\author{
	Changyu Gu$^{1}$ \qquad \qquad Linwei Chen$^{2,4,5}$ \qquad \qquad Lin Gu$^{3}$ \qquad \qquad Ying Fu$^{1*}$ \\
	$^1$Beijing Institute of Technology \quad $^2$University of Hong Kong \quad $^3$Tohoku University\\
	$^4$The Hong Kong University of Science and Technology \\
	$^5$Hong Kong Generative AI Research and Development Center \\
	{\tt\small guchangyu@bit.edu.cn\quad chenlinwei.ai@gmail.com \quad lin@tohoku.ac.jp \quad fuying@bit.edu.cn}
}
\begin{document}
 \maketitle
 \begin{abstract}

In remote sensing rotated object detection, mainstream methods suffer from two bottlenecks, directional incoherence at detector neck and task conflict at detecting head. Ulitising fourier rotation equivariance,  we introduce \textbf{Fourier Angle Alignment}, which analyses angle information through frequency spectrum and aligns the main direction to a certain orientation. Then we propose two plug and play modules : \textbf{FAAFusion} and \textbf{FAA Head}. FAAFusion works at the detector neck, aligning the main direction of higher-level features to the lower-level features and then fusing them. FAA Head serves as a new detection head, which pre-aligns RoI features to a canonical angle and adds them to the original features before classification and regression. Experiments on DOTA-v1.0, DOTA-v1.5 and HRSC2016 show that our method can greatly improve previous work. Particularly, our method achieves new state-of-the-art results of 78.72\% mAP on DOTA-v1.0 and 72.28\% mAP on DOTA-v1.5 datasets with single scale training and testing, validating the efficacy of our approach in remote sensing object detection. The code is made publicly available at \href{https://github.com/gcy0423/Fourier-Angle-Alignment}{https://github.com/gcy0423/Fourier-Angle-Alignment}.

\end{abstract}    
 \section{Introduction}
\label{sec:intro}

Compared to general object detection, objects in remote sensing, such as ships, airplanes, and vehicles, are often oriented in arbitrary directions \cite{dota,zongshu1,zongshu2}. For rotated object detection (ROD),  Oriented Bounding Box (OBB), which simultaneously predicts both the class and the orientation of each object \cite{oob1,oob2}, is a popular solution.

\begin{figure}[t]
  \centering
  \includegraphics[width=1\linewidth]{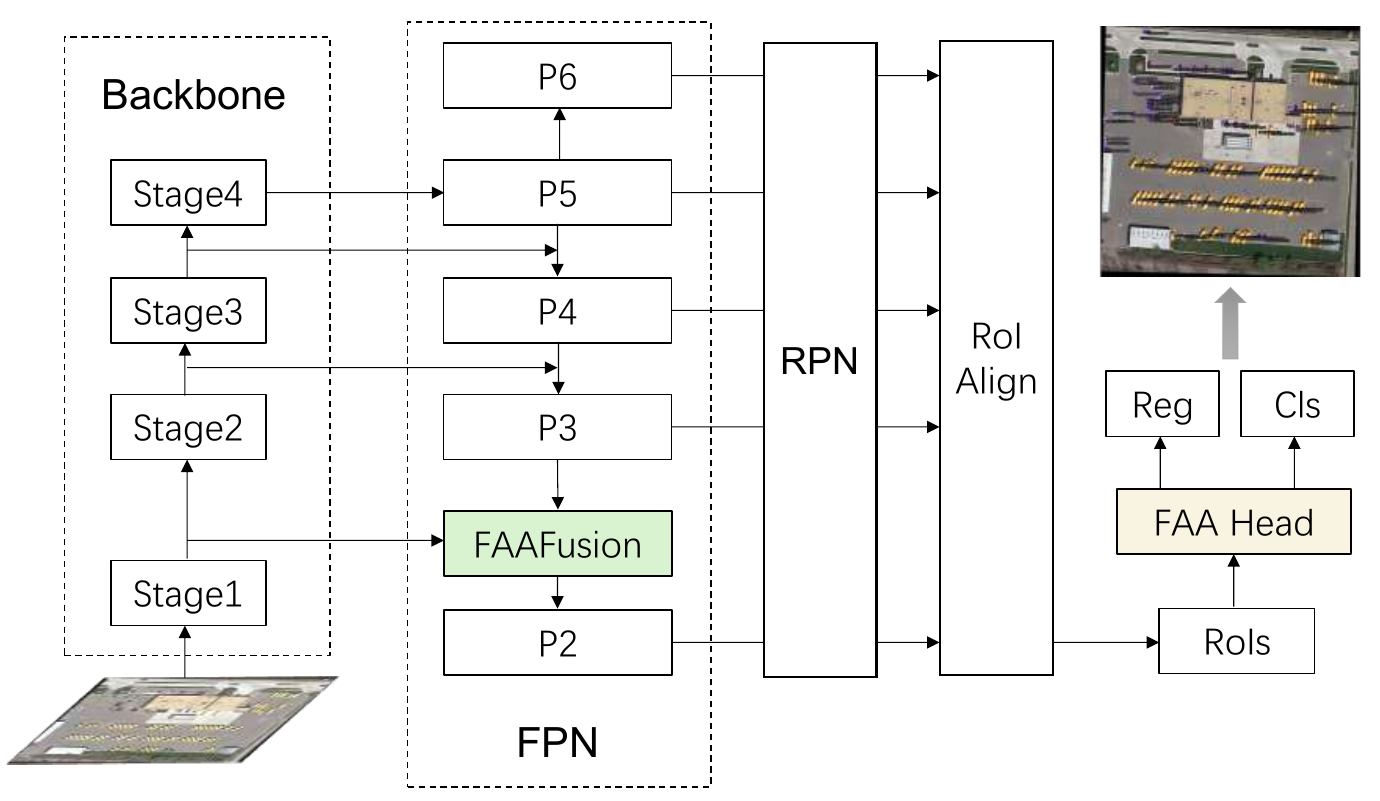}
   \caption{The position of FAAFusion and FAA Head.}
   \label{fig:pipeline}
\end{figure}

Recent progress in ROD focuses on designing rotation-sensitive convolutions \cite{arc,gra,conv1,zhu2025trim}, modifying the backbone \cite{lsknet,pkinet,striprcnn} or detection neck \cite{bifpn,panet,neck1,neck2,neck3}, or optimizing angle regression losses \cite{angle1,angle2,gwd,kld}. In this work, we propose  \textit{Fourier Angle Alignment (FAA)} that leverages frequency domain to  estimate and align object orientation to enhance existing ROD methods. As shown in Figure \ref{fig:pipeline}, our plug-and-play modules can be applied on the main-stream ROD architectures to deal with two common bottlenecks: \textit{directional incoherence in the neck} and    \textit{task conflict in the detection head}.

\begin{figure}[t]
  \centering
  \begin{subfigure}[b]{0.32\linewidth}
    \centering
    \includegraphics[width=\linewidth]{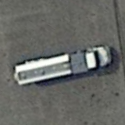} 
    \caption{Original picture}
  \end{subfigure}%
  \hfill
  \begin{subfigure}[b]{0.32\linewidth}
    \centering
    \includegraphics[width=\linewidth]{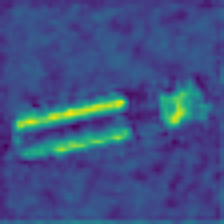} 
    \caption{Low-level feature}
  \end{subfigure}%
  \hfill
  \begin{subfigure}[b]{0.32\linewidth}
    \centering
    \includegraphics[width=\linewidth]{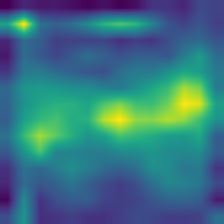} 
    \caption{High-level feature}
  \end{subfigure}

  \caption{Directional incoherence. High-level feature (c) has strong semantics and large receptive fields but captures the coarse orientation. For example, from (c) we can only infer that the object lies horizontally. Yet many down-samplings blur spatial details, so the orientation signal is low-frequency and vague. Low-level feature (b) keeps rich edges, corners, and textures, giving clear and high-frequency orientation cues.}
  \label{fig:threepics}
\end{figure}

When giving features from different levels, as shown in Figure \ref{fig:threepics}, traditional FPN \cite{fpn} fuses them using simple element-wise addition or concatenation. However, on the one hand, features from high levels commonly have strong semantics but blurred spatial details, thus making the orientation coarse and low-frequency \cite{chen2024freqfusion}. On the other hand, features from low levels have sharp edges and textures, so they carry precise, high-frequency direction cues. When these two types of features are directly fused, their orientation signals often conflict. This mismatch creates noise in the fused features, degrading angle prediction accuracy.

\begin{figure}[t]
  \centering
  \includegraphics[width=1\linewidth]{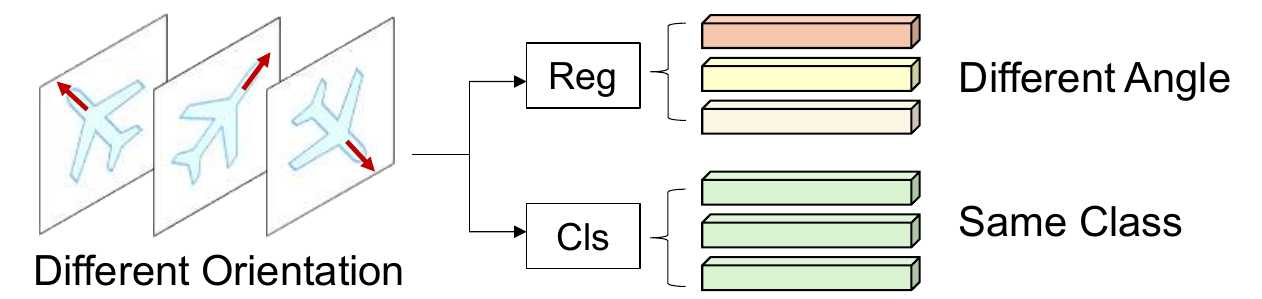}
   \caption{Task conflict in the detection head. As the orientation varies, the angle predictions are expected to be different while the class predictions are expected to be the same.}
   \label{fig:taskconflict}
\end{figure}

Since the same RoI feature must serve two different tasks, classification and angle regression often conflict. As shown in Figure \ref{fig:taskconflict}, classification needs rotation-invariant features, for a plane should always be recognized as a plane regardless of its orientation. Regression, however, needs rotation-sensitive features, where the feature must change when the object rotates. Forcing a single feature to satisfy both goals leads to a compromise: the feature becomes neither fully invariant nor fully sensitive. This limits the model’s ability to accurately predict both class and angle.

Our FAA comprises two lightweight modules:
\textbf{FAAFusion} and \textbf{FAA Head}. Embedded in the FPN  \cite{fpn}, FAAFusion first estimates the dominant orientation from low-level features using Fourier analysis and then rotates the high-level features to align them with the low-level features before fusion. This ensures consistent orientation across scales. FAA Head serves as a new detection head. For each RoI features, it estimates the dominant orientation in the frequency domain and rotates the RoI feature to a canonical orientation. Then it adds the rotated feature back to the original one. The aligned feature will benefit the task of classification, while the original feature is helpful for location and orientation regression.

We evaluate our method on three challenging benchmark datasets: DOTA-v1.0 \cite{dota}, DOTA-v1.5 \cite{dota}, and HRSC2016 \cite{hrsc2016}. We select three representative backbones of ResNet50 \cite{resnet}, LSKNet-S \cite{lsknet} and StripNet-S \cite{striprcnn} within the Oriented R-CNN \cite{orcnn} framework. Our method improves the mAP of these state-of-the-art methods by 0.68\%, 1.00\% and 0.63\% on DOTA-v1.0 \cite{dota}, 0.37\%, 2.02\% and 1.73\% on DOTA-v1.5 \cite{dota}, 2.17\%, 1.81\% and 1.23\% on HRSC2016 \cite{hrsc2016}, proving that frequency-domain orientation estimation is a powerful and practical tool for rotated object detection.

Our main contributions could be summarized as:  
 \begin{itemize}
     \item We propose Fourier Angle Alignment, a plug-and-play framework that leverages frequency-domain analysis to estimate and align object orientations, thereby enhancing existing rotated object detection methods.
    \item We address directional incoherence at detector neck with the FAAFusion module, which explicitly aligns local features by orientation to ensure directional semantic consistency across scales.
    \item  The FAA Head module replaces detection head and pre-aligns RoI features to a canonical orientation and fuses them with the original features, effectively mitigating the conflict between classification and angle regression.
\end{itemize}

 \section{Related Works}
\label{sec:relatedworks}

This section briefly reviews the most relevant studies and methods that have advanced the field, eagerly hoping to lay the foundation for the contributions of this work.

\subsection{Remote Sensing Object Detection}
Given that objects in remote sensing imagery can be oriented in any direction, object detectors must extend beyond the standard outputs of general detection to additionally encode the orientation of the bounding box. Mainstream methods could be summarized as follows. Some studies pay attention to the improvement of the backbone network, replacing the normal convolution block in ResNet \cite{resnet} with well-designed blocks. ARC \cite{arc} introduces an adaptive rotation convolution module which applies multiple groups of rotated kernels to extract features of different orientations. GRA \cite{gra} uses group-wise rotating and group-wise attention to relieve the noises produced by the rotated convolution kernels in undesired direction.  Some studies focus on building new backbones. As the first method introducing RE-Nets into oriented object detection networks, ReDet \cite{redet} proposes ReResNet as a rotation-equivariant backbone to inherently preserve orientation information. LSKNet \cite{lsknet} adaptively captures contextual information of multi-scale and arbitrarily oriented objects in aerial images by introducing a Large Selective Kernel mechanism. PKINet \cite{pkinet} enhances the ability to perceive objects with extreme aspect ratios and varying orientations in aerial images by employing learnable parameterized large convolution kernels. Strip R-CNN \cite{striprcnn} employs a strip convolution structure to efficiently model the geometric characteristics of slender objects in aerial images, improving detection performance for high aspect-ratio rotated targets. With the strong feature extraction abilities of their backbone networks, these approaches have vitally enhanced detecting performance.

Other studies concentrate on the neck and detecting head design to better accommodate oriented bounding box (OBB) prediction. While general-purpose necks like PANet \cite{panet} and BiFPN \cite{bifpn} enhance cross-scale feature fusion and benefit multi-scale detection, they do not explicitly model orientation. In contrast, top-tier methods integrate orientation awareness into the detection head or RoI processing stage: RoI Transformer \cite{roitrans} learns to transform axis-aligned RoI features into oriented ones via a lightweight spatial transformer, effectively aligning features before regression. Oriented R-CNN \cite{orcnn} adopts a two-stage head that decouples angle prediction from box refinement, achieving high accuracy with low computational overhead. S2A-Net \cite{s2anet} introduces an anchor-free single-stage head with an AlignConv module that dynamically samples features according to predicted angles. Crucially, significant advances have been made in OBB parameterization and regression losses: GWD \cite{gwd} models OBBs as 2D Gaussian distributions and uses Wasserstein Distance as the loss, while KLD  \cite{kld} replaces it with Kullback–Leibler Divergence to ensure scale invariance and improve training stability. Despite these impressive results, these approaches primarily operate on spatial-domain features and fail to explicitly encode or learn rotational information directly from the feature representation itself.

\subsection{Frequency-domain Analysis}

Frequency-domain analysis has a long and rich history in computer vision. Early approaches \cite{fhog,kai1,rihog,cje2} primarily used Fourier properties to design rotation-invariant handcrafted feature extractors, then used these features for downstream tasks. In recent years, with the rise of deep learning, researchers \cite{song1,zhang2025unaligned,song2,zhang2026supervise,cje1} have begun to explicitly integrate Fourier transforms into neural network architectures to exploit their global receptive fields and inherent properties under geometric transformations. For instance, FcaNet \cite{fcanet} replaces global average pooling with a frequency-channel attention mechanism, modeling long-range dependencies in the spectral domain to significantly boost performance in image classification and object detection. More recently, a series of works\cite{chen2024fdc,cje0} have underscored the critical role of explicit frequency modeling in dense prediction. Spatial Frequency Modulation \cite{sfm} mitigates aliasing during downsampling via a modulation-demodulation scheme to preserve high-frequency details. FreqFusion \cite{chen2024freqfusion} decomposes features into low- and high-frequency components to jointly enhance intra-class consistency and boundary sharpness. Frequency-Adaptive Dilated Convolution \cite{chen2024fadc} dynamically adjusts dilation rates and kernel spectral responses based on local frequency content to balance receptive field size and effective bandwidth. And modules like FDAM \cite{chen2025fda} counteract the low-pass bias of Vision Transformers through high-frequency attention compensation. Collectively, these advances demonstrate that explicit spectral modeling has established a powerful foundation for improving the performance of feature extraction, thus providing a new view of oriented object detection.

\section{Methods}
\label{sec:methods}
In this section, we first introduce our formulation and motivation. Next we describe the detailed method of our Fourier Angle Alignment. Then we will present the applications of FAA at different locations of the detector, FAAFusion and FAA Head, which will be detailed in \ref{sec:faafusion} and \ref{sec:faahead}.

\begin{figure}[t]
\centering
\begin{minipage}[t]{0.49\linewidth}
  \centering
  \includegraphics[width=0.32\linewidth]{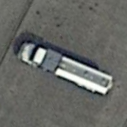}\hfill
  \includegraphics[width=0.32\linewidth]{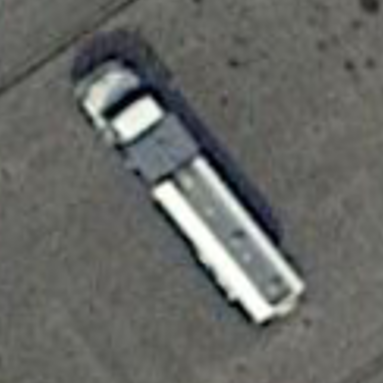}\hfill
  \includegraphics[width=0.32\linewidth]{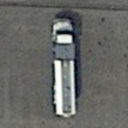}\\[0.5pt]
  \includegraphics[width=0.32\linewidth]{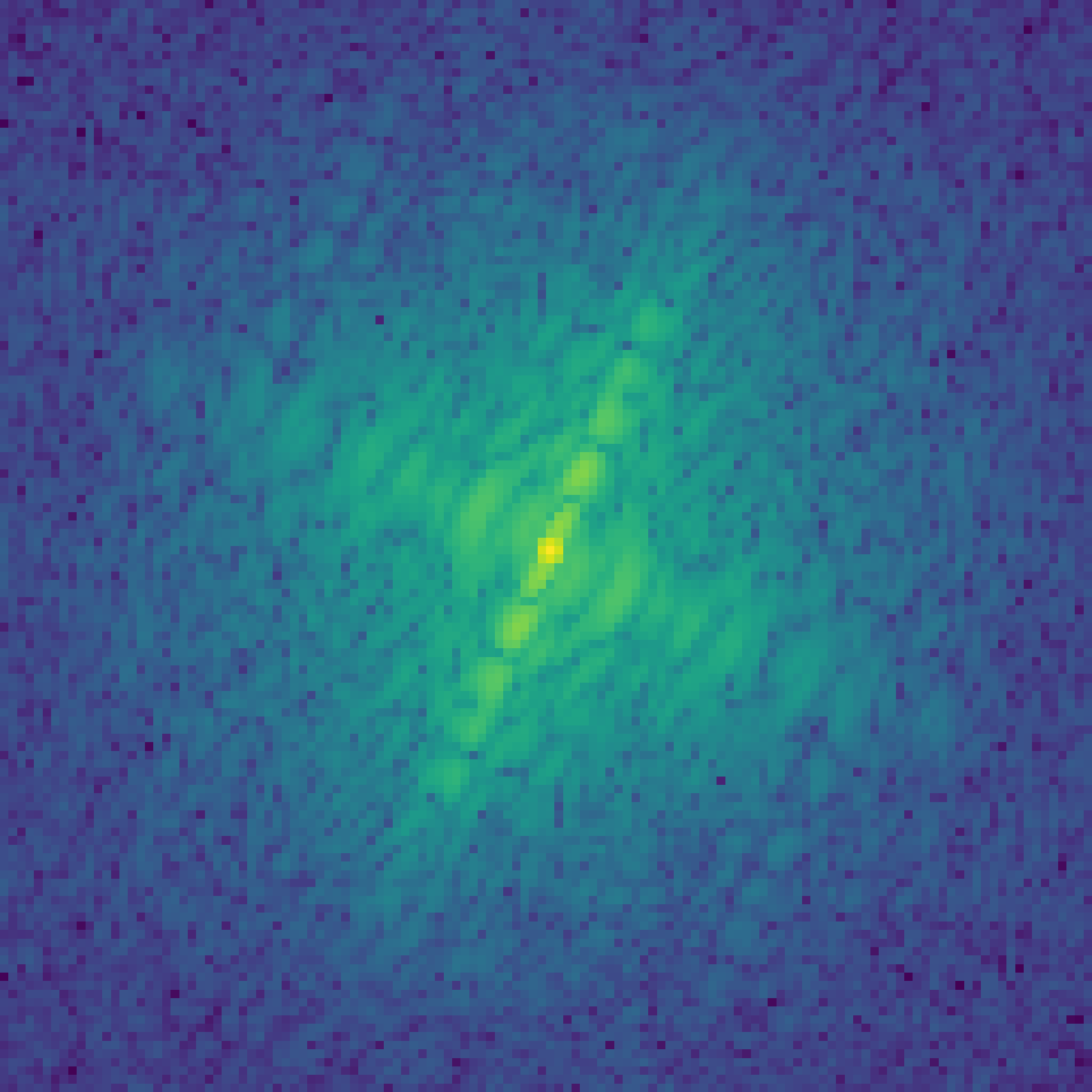}\hfill
  \includegraphics[width=0.32\linewidth]{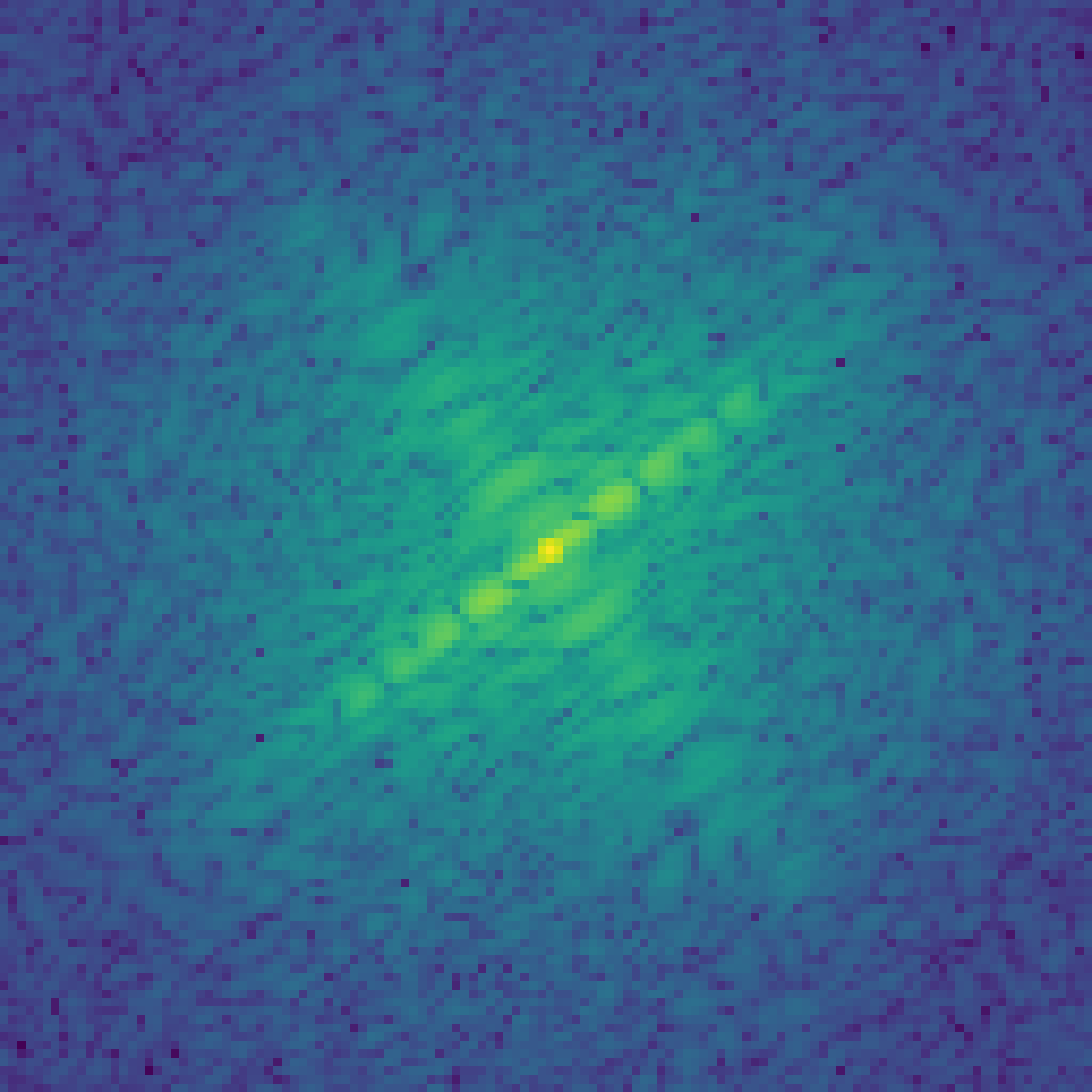}\hfill
  \includegraphics[width=0.32\linewidth]{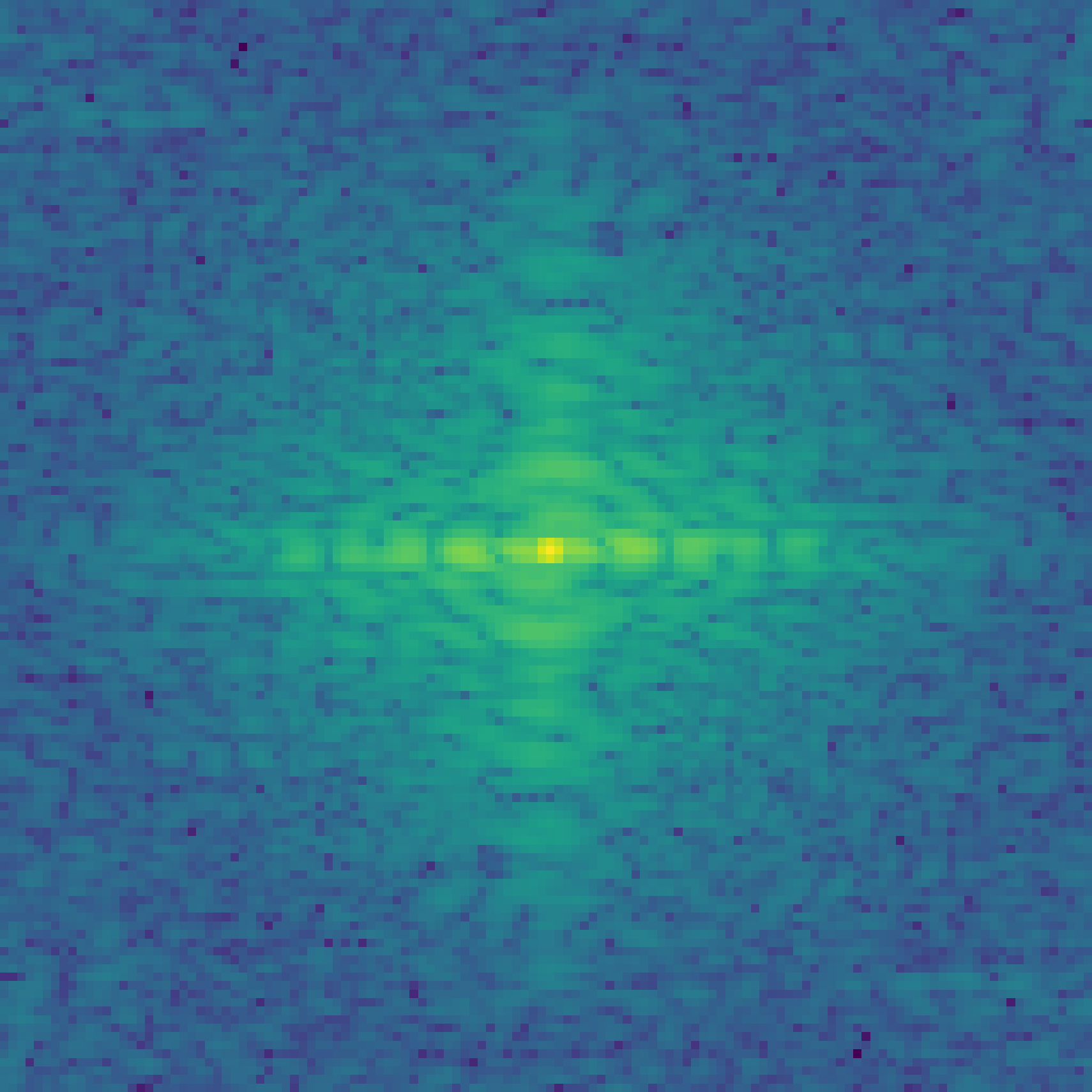}\\[0.5pt]
  \includegraphics[width=0.32\linewidth]{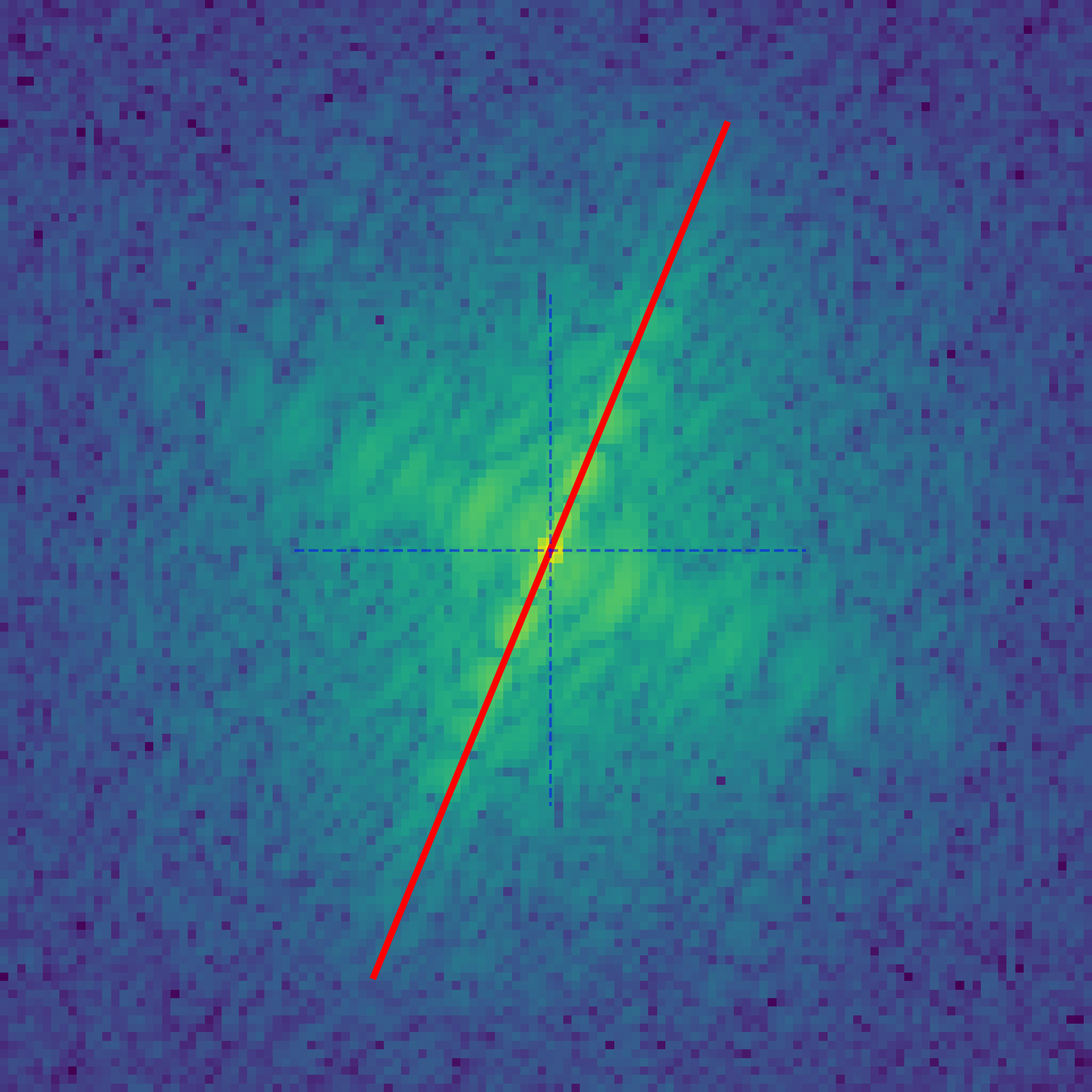}\hfill
  \includegraphics[width=0.32\linewidth]{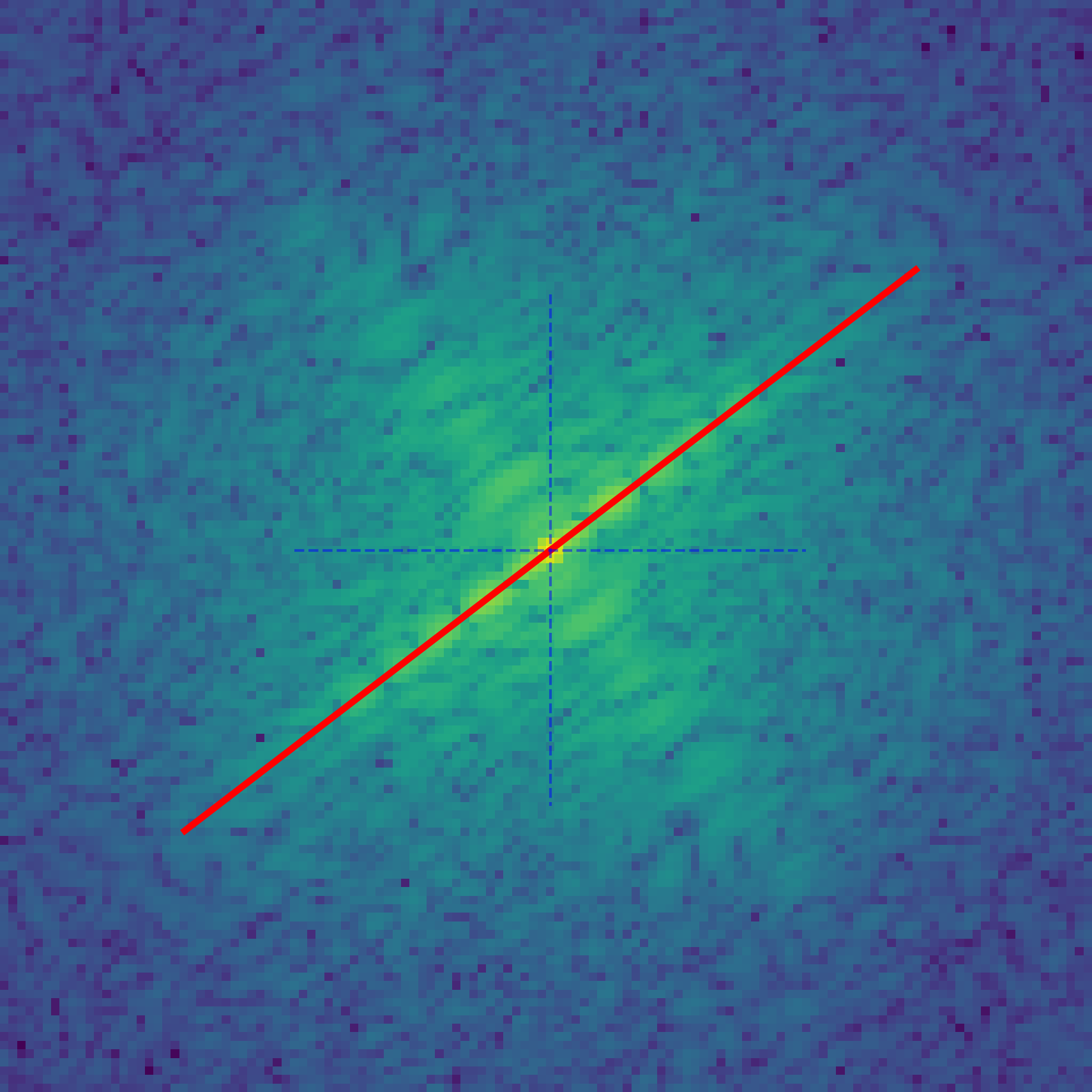}\hfill
  \includegraphics[width=0.32\linewidth]{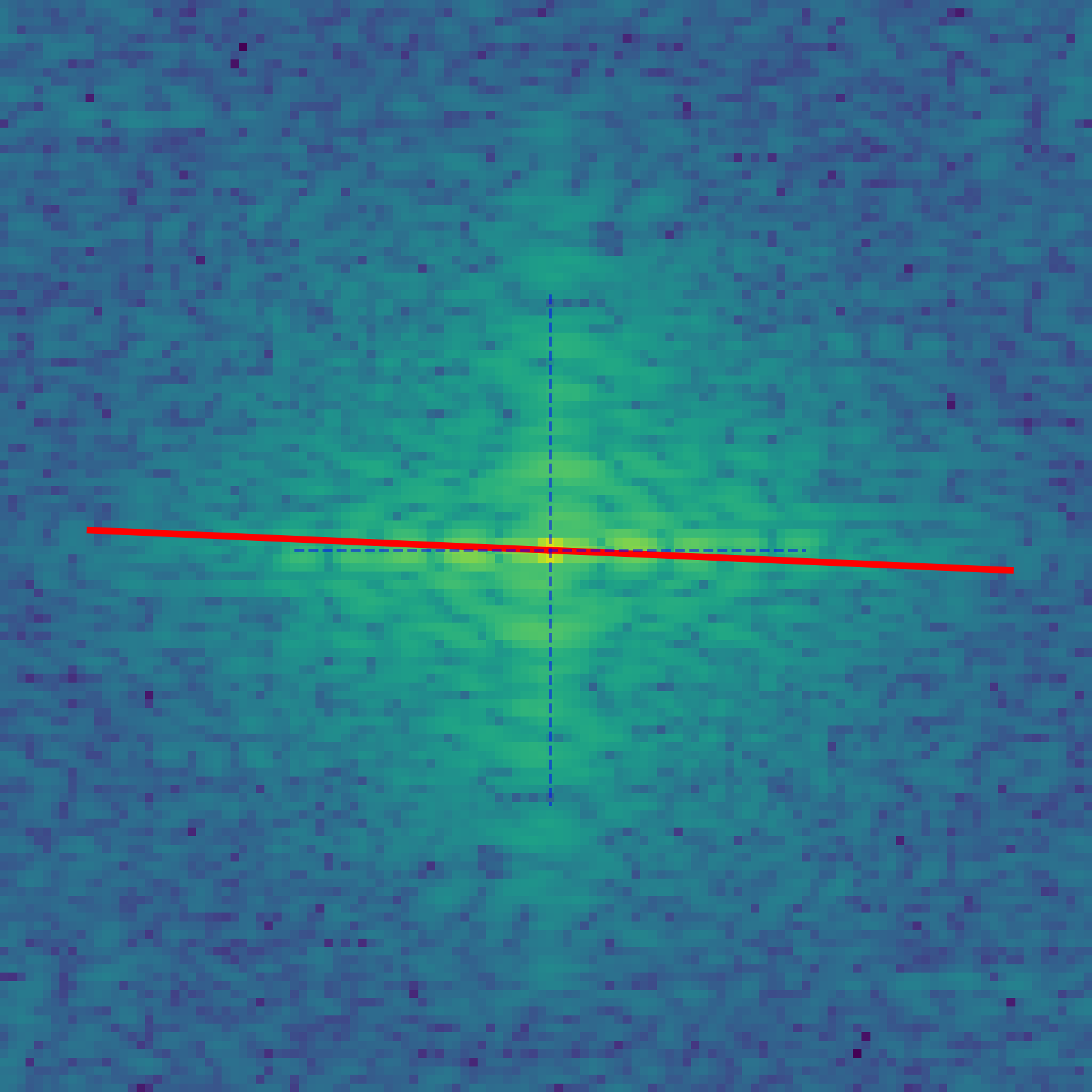}
  \\[2pt]  
  \textbf{\footnotesize (a) Rotated Object}
\end{minipage}\hfill
\begin{minipage}[t]{0.49\linewidth}
  \centering
  \includegraphics[width=0.32\linewidth]{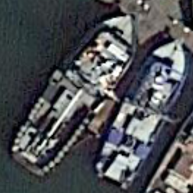}\hfill
  \includegraphics[width=0.32\linewidth]{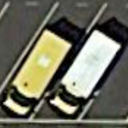}\hfill
  \includegraphics[width=0.32\linewidth]{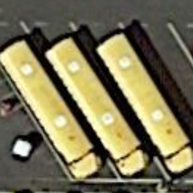}\\[0.5pt]
  \includegraphics[width=0.32\linewidth]{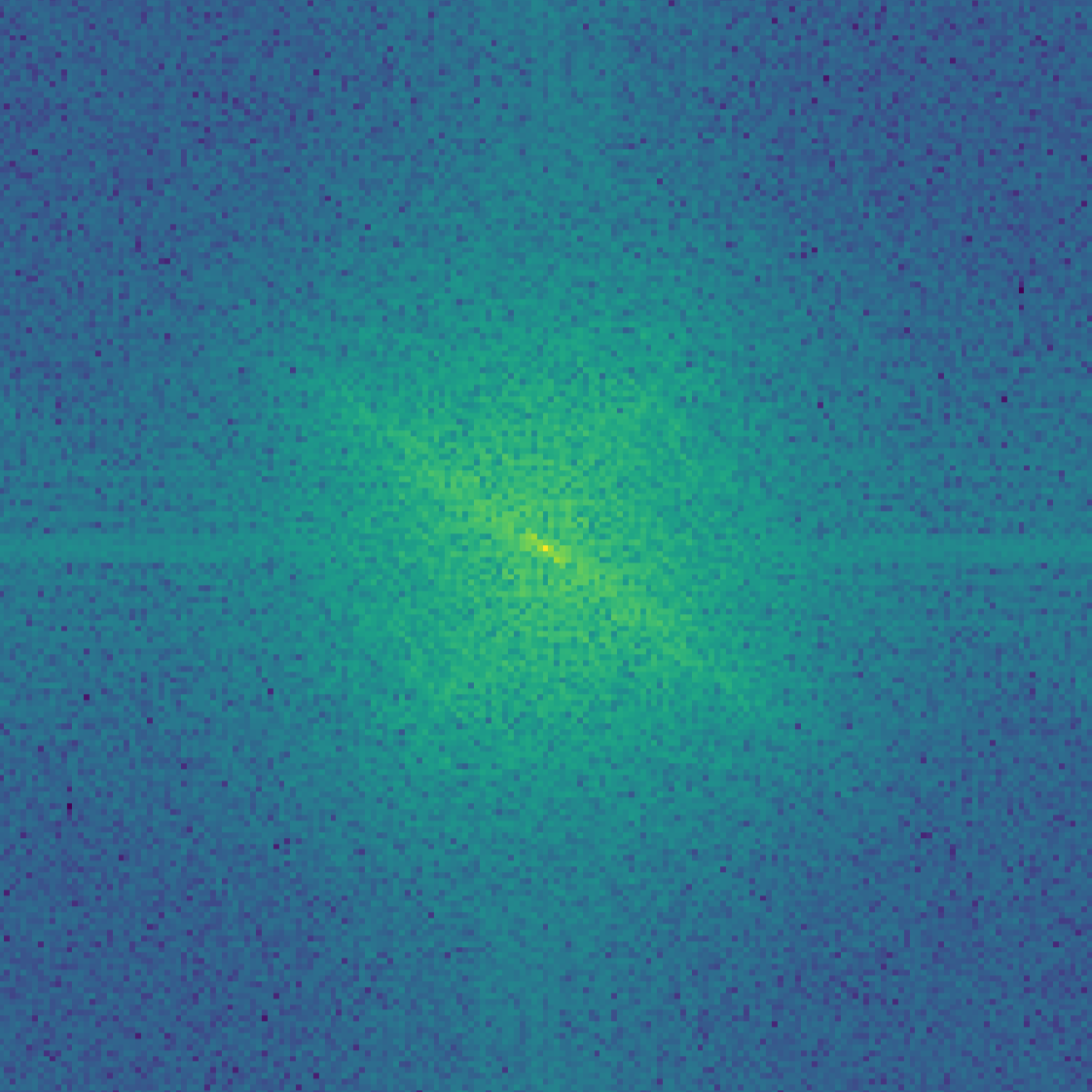}\hfill
  \includegraphics[width=0.32\linewidth]{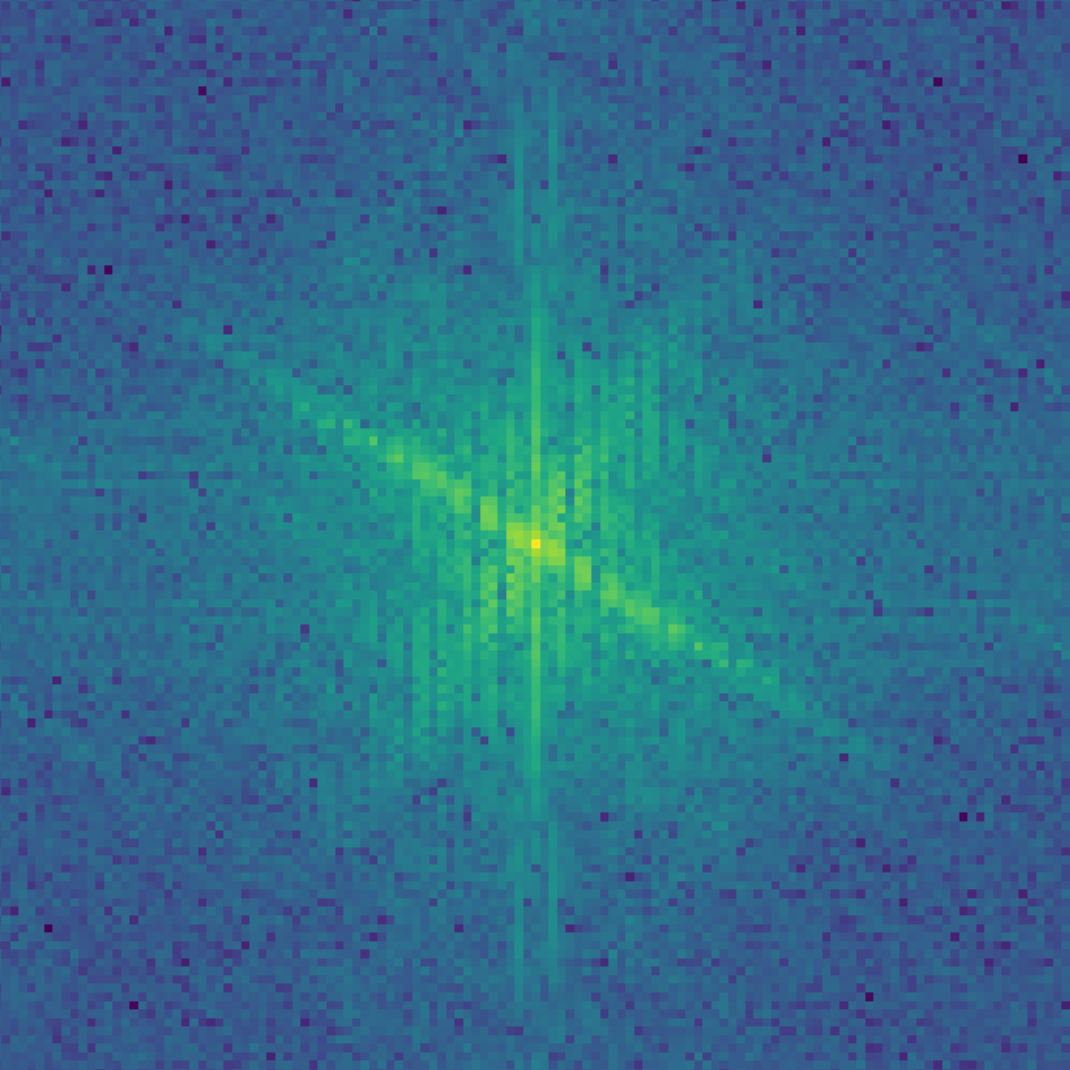}\hfill
  \includegraphics[width=0.32\linewidth]{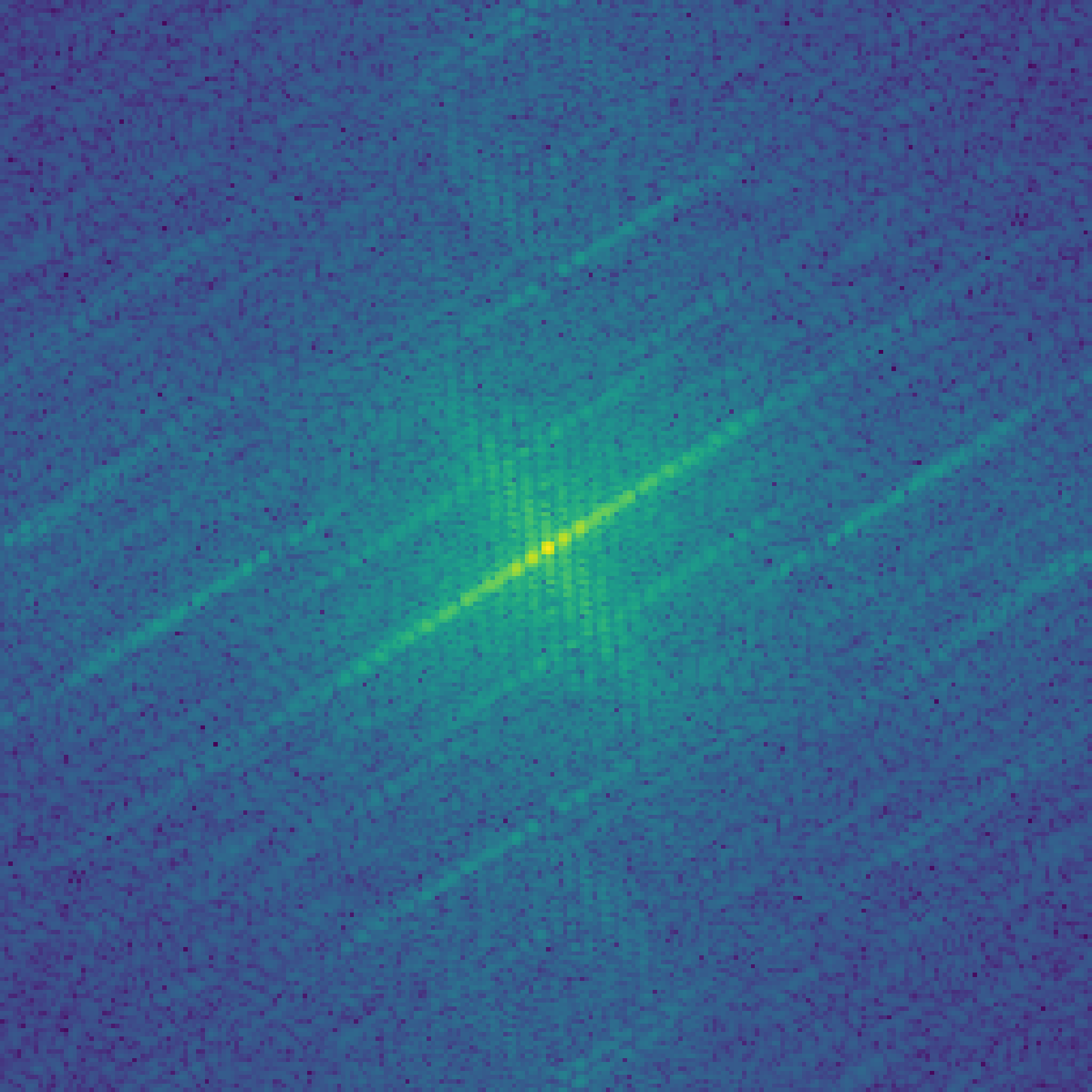}\\[0.5pt]
  \includegraphics[width=0.32\linewidth]{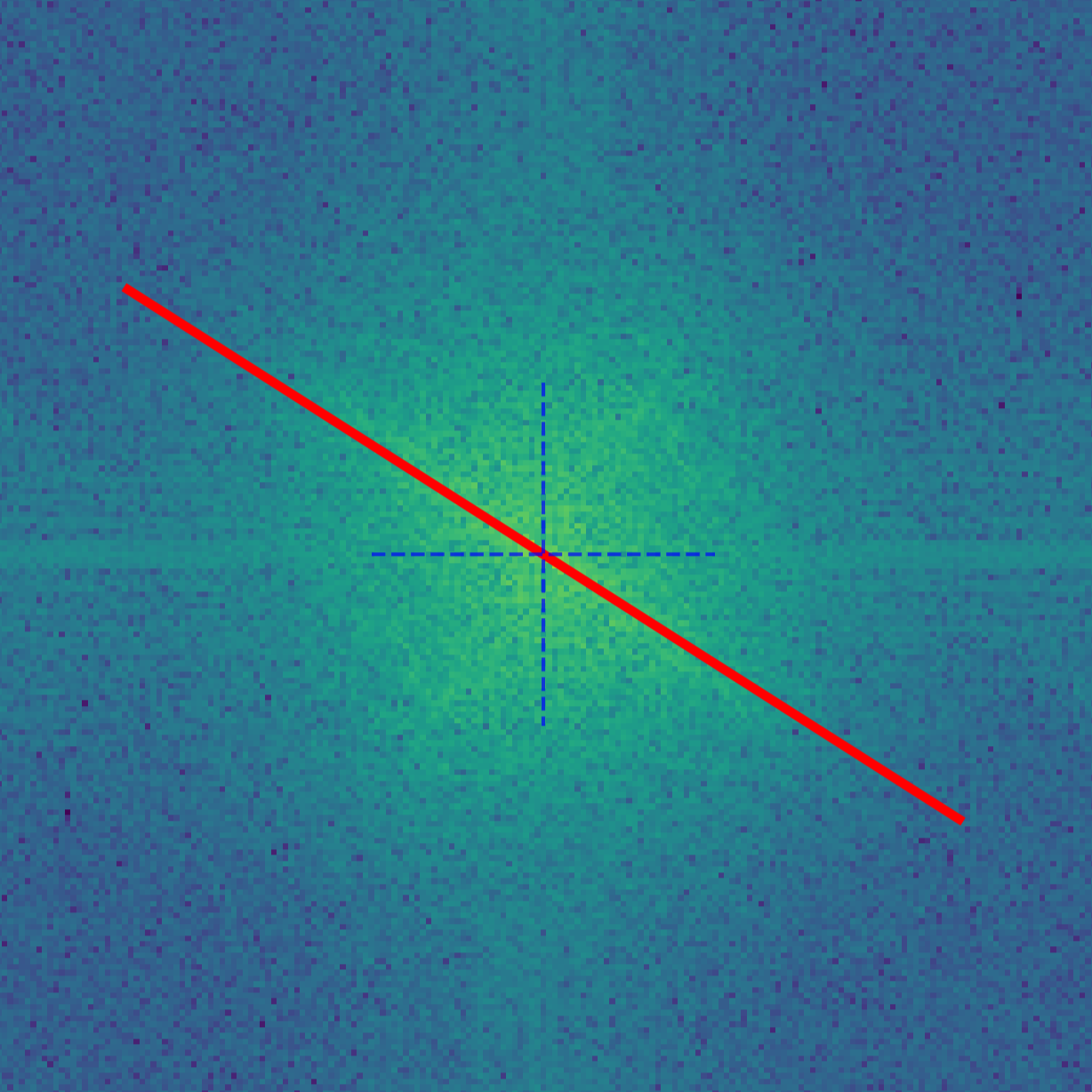}\hfill
  \includegraphics[width=0.32\linewidth]{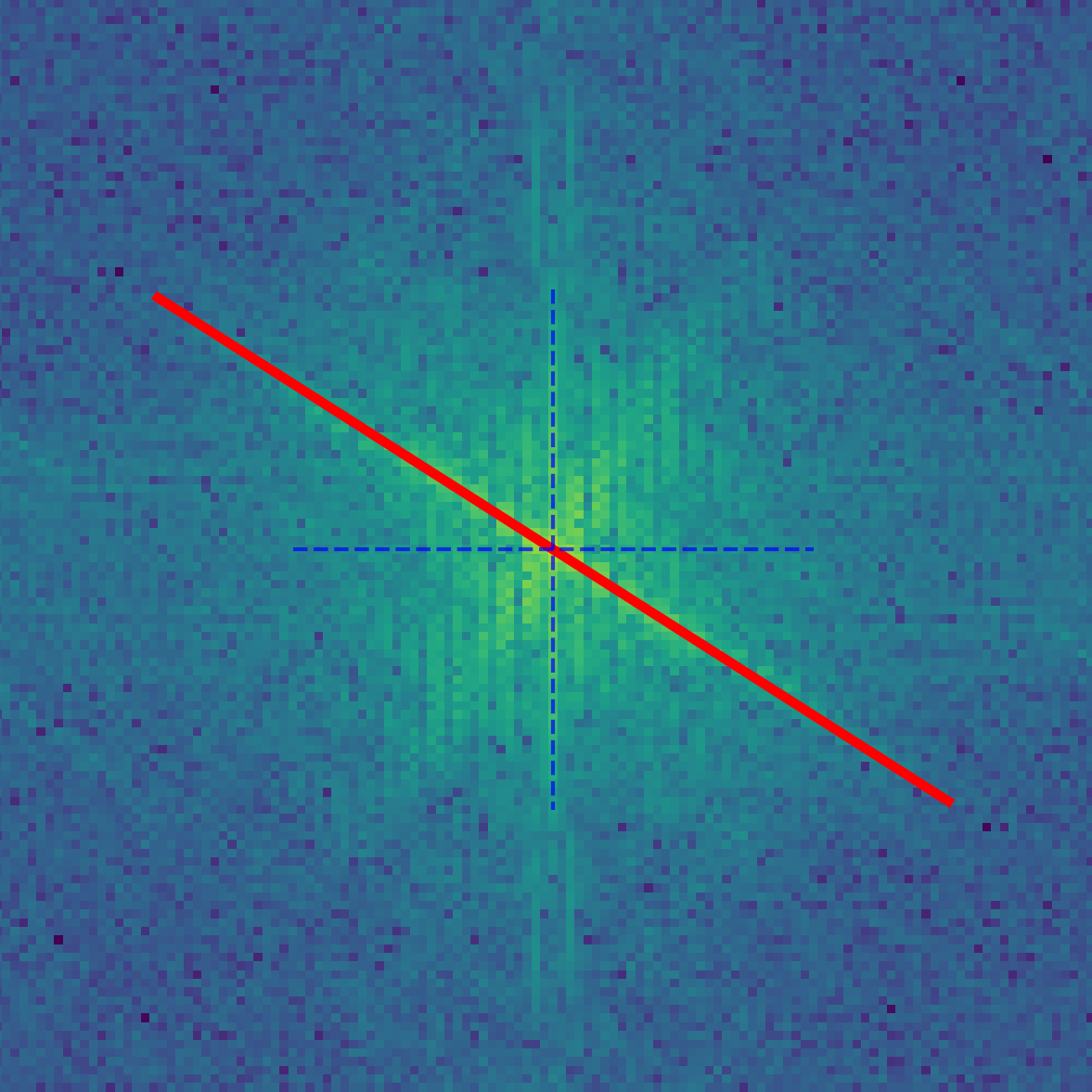}\hfill
  \includegraphics[width=0.32\linewidth]{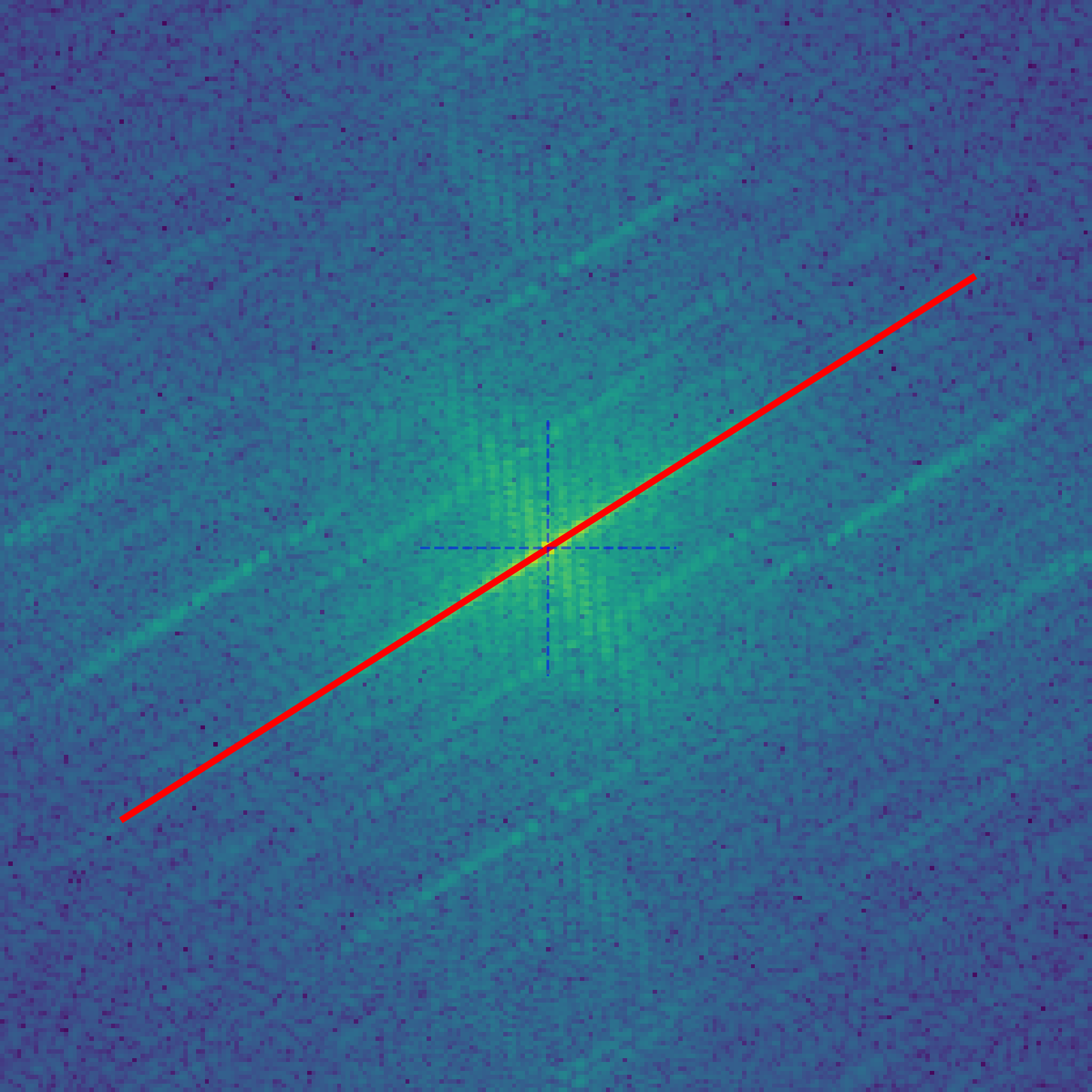}
  \\[2pt]  
  \textbf{\footnotesize (b) Rectangle-like Object}
\end{minipage}

\begin{subfigure}{0.49\linewidth}\phantomsubcaption\label{fig:spectrum:left}\end{subfigure}%
\begin{subfigure}{0.49\linewidth}\phantomsubcaption\label{fig:spectrum:right}\end{subfigure}%

\caption{The top row shows the target images in the spatial domain. The second row gives their matching frequency spectra. In the bottom row, red lines point out the main energy directions in each spectrum.
\textbf{(a)} If the spatial picture is rotated, its spectrum distribution will be rotated as well.  
\textbf{(b)} The main spectral direction of one rectangle-like object is perpendicular to the direction of its major axis.}
\label{fig:spectrum}
\end{figure}

\subsection{Formulation and Motivation}

\noindent\textbf{Oriented Object Detection.}
Let the input image be $ \textbf{I}(\textbf{x}) $, $ \textbf{x}=(x,y)^\intercal $.
For a horizontal object detector $ \mathcal{D}_{\text{HBB}} $, the output is a set of axis-aligned bounding boxes with class information:
\begin{equation}
\mathcal{D}_{\text{HBB}}(\textbf{I}) = \{(x_i, y_i, w_i, h_i, c_i)\},
\end{equation}
where $(x_i,y_i)$ is the box center, $h_i$ and $w_i$ stand for the hight and width of the box, and $c_i$ is the class of the object in the box. 

For an oriented object detector $ \mathcal{D}_{\text{OBB}} $, the output is a set of rotated bounding boxes that include an orientation angle:
\begin{equation}
\mathcal{D}_{\text{OBB}}(\textbf{I}) = \{(x_i, y_i, w_i, h_i, \theta_i, c_i)\},
\end{equation}
where $ x_i, y_i,w_i, h_i,c_i $ are the same as the HBB, and $ \theta_i$ is the rotation angle.

\noindent\textbf{Fourier Rotation Equivariance.}
\label{para:fourierRE}
Let $ \textbf{I}(\textbf{x}) $ denote a signal in the spatial domain. Its frequent spectrum is
\begin{equation}
    \textbf{F}(\boldsymbol{\omega}) = \mathcal{F}\{\textbf{I}(\textbf{x})\},
\end{equation}
where $\boldsymbol{\omega}=(u,v)^\intercal$ and $\mathcal{F}$ means Fourier Transform. Next, consider rotating $ \textbf{I} $ counterclockwise by an angle $ \phi $ around the origin, and we can know that the rotated image will be $ \textbf{I}_\phi(\mathbf{x}) = \textbf{I}(\mathbf{R}_{-\phi} \mathbf{x}) $, where $ \mathbf{R}_\phi $ is the 2D rotation matrix:
\begin{equation}
    \mathbf{R}_\phi =
    \begin{bmatrix}
        \cos\phi & -\sin\phi \\
        \sin\phi & \cos\phi
    \end{bmatrix}.
\end{equation}

Then, suppose the rotated spectrum is $ \textbf{F}_\phi(\boldsymbol{\omega})$. 
By a change of variables, it can finally be inferred that:
\begin{equation}
    \textbf{F}_\phi(\boldsymbol{\omega}) = \mathcal{F}\{\textbf{I}(\mathbf{R}_{-\phi}\textbf{x})\} = \mathcal{F}\{( \textbf{I}_\phi(\mathbf{x}) )\},
\end{equation}
which means if a spatial signal is rotated by $\phi$, its frequent spectrum will also be rotated by $\phi$, as shown in Figure \ref{fig:spectrum:left}.

\noindent\textbf{Spectral alignment of rectangles.}
Suppose $ \textbf{I}_\text{rec}(\mathbf{x}) $ denote a 2D rectangular function in the spatial domain and $ \textbf{x} = (x, y)^\intercal $ Specifically,
\begin{equation}
    \textbf{I}_\text{rec}(\mathbf{x}) = 
    \begin{cases}
        1, & |x| \leq a,\; |y| \leq b, \\
        0, & \text{otherwise},
    \end{cases}
\end{equation}
with $ 0 < b \leq a $, so that the major axis of the rectangle is aligned with the $x$-axis. Its frequent spectrum is
\begin{equation}
    \textbf{F}_\text{rec}(\boldsymbol{\omega}) = \mathcal{F}\{\textbf{I}_\text{rec}(\mathbf{x})\},
\end{equation}
where $ \boldsymbol{\omega} = (u, v)^\intercal $ denotes the frequency-domain coordinates. By separability,
\begin{equation}
    \textbf{F}_\text{rec}(\boldsymbol{\omega}) = 4ab \cdot \operatorname{sinc}(2a u) \cdot \operatorname{sinc}(2b v),
\end{equation}
where $ \operatorname{sinc}(z) = \sin(\pi z)/(\pi z) $, and the the power spectrum of the rectangle is
\begin{equation}
|\textbf{F}_\text{rec}(\boldsymbol{\omega})|^2 \propto \operatorname{sinc}^2(2a u) \operatorname{sinc}^2(2b v).
\end{equation}

Since $ a > b $, the main lobe of $ \operatorname{sinc}(2a u) $ is narrower than that of $ \operatorname{sinc}(2b v) $. Hence, spectral energy decays more slowly along the $v$-axis than along the $u$-axis. Consequently, in the high-frequency regime, the dominant spectral energy lies along the $v$-axis, which is perpendicular to the major axis of the rectangle, as shown in Figure \ref{fig:spectrum:right}.

\noindent\textbf{Motivations.}
These observations reveal a fundamental property: object orientation can be reliably estimated from the frequency domain. This offers a direct and principled way to handle orientation, which is largely ignored by current detectors that rely only on spatial features.

Specifically, in the neck, low-level features contain sharp edges and textures that encode precise directional cues, while high-level features carry strong semantics but have blurred orientation. Simply adding these features together mixes directionally inconsistent signals, harming angle prediction. Instead, we can use the orientation estimated from low-level features in the frequency domain to explicitly align the high-level features before fusion, ensuring directional consistency across scales.

Similarly, in the detection head, the same RoI feature is expected to support both classification that benefits from rotation-invariant features and angle regression that needs rotation-sensitive features. This creates an inherent conflict. By estimating the dominant orientation of the RoI in the frequency domain, we can rotate the feature to a canonical pose. The resulting aligned feature becomes nearly rotation-invariant for classification, while the original RoI features is helpful for regression, which could effectively decouple the two conflicted tasks.

\begin{figure*}[t]
  \centering
  \includegraphics[width=1\linewidth]{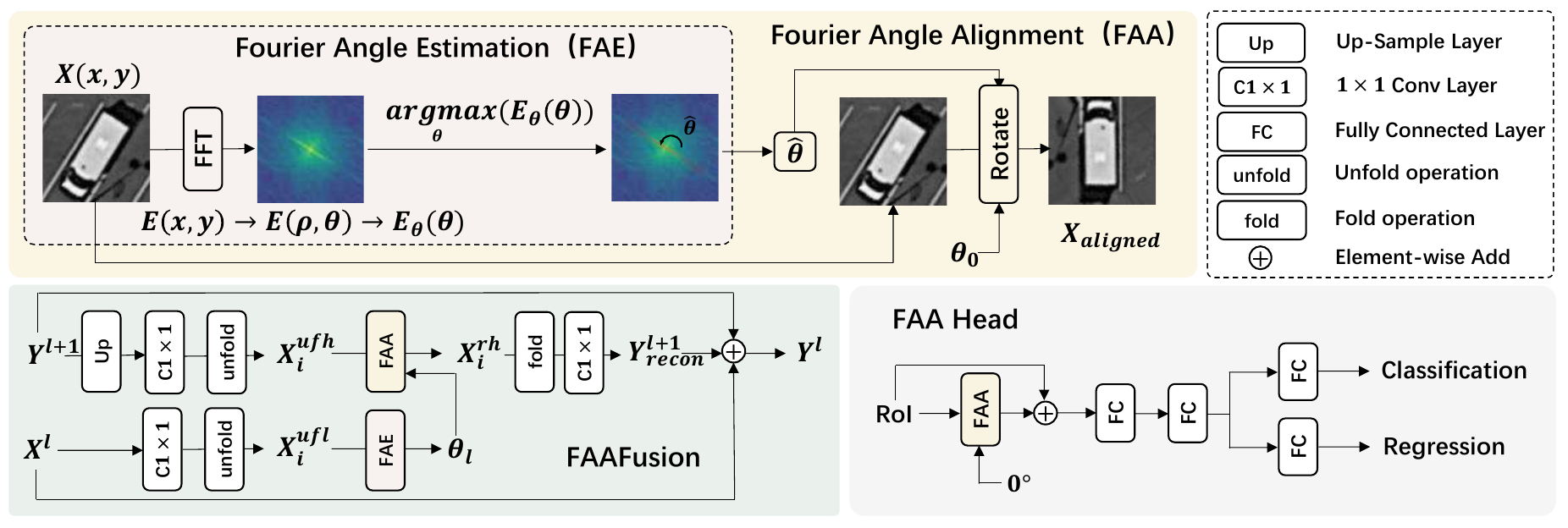}
   \caption{Structures of FAA,  FAAFusion and FAA Head. For FAA, it can be divided into two steps. First, we estimate the main direction in frequent domain. And then FAA accepts an external angle and aligns the main direction with it. For FAAFusion, we use the main direction of the lower local feature as the external angle and align the higher local feature with it. For FAA Head, it aligns the main direction of RoI features with $ 0\degree $ to obtain the rotation invariant features.}
   \label{fig:mainmethod}
\end{figure*}

\subsection{Fourier Angle Alignment}

We propose Fourier Angle Alignment (FAA), with the main progress shown in Figure \ref{fig:mainmethod}. Generally speaking, FAA can be divided into two steps: Fourier Angle Estimation and Angle Alignment.

\noindent\textbf{Fourier Angle Estimation.}
With a square feature map $\mathbf{X} \in \mathbb{R}^{H \times H}$ as input, the FAA estimates its dominant orientation and performs rotational alignment. To begin with, FAA applis a 2D discrete Fourier transform to the input feature map to obtain its spectral representation:
\begin{align}
    \mathbf{F} = \mathcal{F}(\mathbf{X}),
\end{align}
where $\mathcal{F}(\cdot)$ denotes the 2D Fourier transform operator. To facilitate subsequent polar coordinate conversion, FFA shifts the zero-frequency component to the center of the spectrum, which is achieved by multiplying with a phase factor:
\begin{equation}
    \begin{aligned}
    \mathbf{F}_c(u, v) &= \mathbf{F}(u, v) \cdot (-1)^{u + v}, \\
    u, v &= 0, 1, \dots, H-1.
    \end{aligned}
\end{equation}

With the intention of estimating the main direction more conveniently, FAA converts the centered spectrum $\mathbf{F}_c$ from Cartesian coordinates $(u, v)$ to polar coordinates $(\rho, \theta)$, and then computes the corresponding energy spectrum:
\begin{align}
    E(\rho, \theta) = \left| \mathbf{F}_c\big(u(\rho, \theta), v(\rho, \theta)\big) \right|^2,
\end{align}
where $u(\rho, \theta) = \frac{H}{2} + \rho \cos\theta$, $v(\rho, \theta) = \frac{H}{2} + \rho \sin\theta$, $\rho \in [0, \rho_{\max}]$, and $\theta \in [0, 2\pi)$.

Based on our prior knowledge, for a rectangular object, its spectral energy is concentrated around specific orientations. To robustly estimate this dominant orientation, we sum the magnitude spectrum in polar coordinates along the radial direction, thereby obtaining a one-dimensional angular energy distribution function $E_{\theta}(\theta)$:
\begin{align}
    E_{\theta}(\theta) = \sum_{\rho} \rho \cdot E(\rho, \theta).
\end{align}

And the estimated main direction of the object is the direction where the $E_{\theta}(\theta)$ reaches maximum. For convenience, we constrain the angular range to $[0, \pi)$.
\begin{equation}
    \begin{aligned}
    \hat{\theta} = \arg\max_{\theta} E_{\theta}(\theta), \\
    \end{aligned}
\end{equation}

\noindent\textbf{Angle Alignment.}
After obtaining the estimated main direction, FAA rotates the original feature map by a specific reference angle to align the main direction with this reference angle:
\begin{align}
    \mathbf{X}_{\text{aligned}} = \mathcal{R}_{\Delta\theta}(\mathbf{X}),
\end{align}
where $\mathcal{R}_{\Delta\theta}(\cdot)$ denotes a 2D rotation operator about the image center. $\Delta\theta = \theta_0 - \hat{\theta}$, where $\theta_0$ is the reference angle.

\subsection{FAAFusion}
\label{sec:faafusion}

FAAFusion replaces the conventional element-wise addition in the FPN \cite{fpn}. Let the high-level semantic feature be $\mathbf{Y}^{l+1} \in \mathbb{R}^{C \times H \times W}$ and the low-level detail feature be $\mathbf{X}^{l} \in \mathbb{R}^{C \times 2H \times 2W}$. 

First, as in the original FPN \cite{fpn}, we upsample the high-level feature $\mathbf{Y}^{l+1}$ to match the spatial resolution of the low-level feature $\mathbf{X}^{l}$.
\begin{align}
\mathbf{Y}^{l+1}_u = \mathrm{Upsample}(\mathbf{Y}^{l+1}).
\end{align}

To reduce computational cost, we apply $1\times1$ convolutions to both $\mathbf{Y}^{l+1}_u$ and $\mathbf{X}^{l}$ to project their channel dimensions to a common intermediate size $C_{mid}$, and then we use the \texttt{unfold} operation to obtain the local features:
\begin{equation}
\begin{aligned}
\{\mathbf{p}^h_i\}_{i=1}^N &= \mathrm{unfold}(\mathrm{Conv}(\mathbf{Y}^{l+1}_u)), \\
\{\mathbf{p}^l_i\}_{i=1}^N &= \mathrm{unfold}( \mathrm{Conv}(\mathbf{X}^{l})),
\end{aligned}
\end{equation}
where $N = 2H\times 2W$ is the total number of spatial locations, and each local feature $\mathbf{p}^h_i, \mathbf{p}^l_i$ corresponds to spatial position $i$. For $i$ from $1$ to $N$, we first estimate the dominant orientation of the lower-level features:
\begin{align}
\theta^l_i = \mathrm{FAE}(\mathbf{p}^l_i).
\end{align}

Then we use $\theta^l_i$ as the aiming orientation to rotate the higher-level features. The rotated higher-level features are
\begin{align}
\mathbf{p}^{rh}_i = \mathrm{FAA}(\mathbf{p}^h_i; \theta^l_i).
\end{align}

After processing all positions $i = 1, \dots, N$, we reconstruct the aligned high-level feature map using the \texttt{fold} operation and apply a $1\times1$ convolution to restore the channel dimension to the output size, resulting in $\mathbf{Y}_{\mathrm{recon}}^{l+1}$:
\begin{align}
\mathbf{Y}_{\mathrm{recon}}^{l+1} = \mathrm{Conv} \left( \mathrm{fold}(\{\mathbf{p}^{rh}_i\}_{i=1}^N) \right).
\end{align}

Finally, the fused feature at level $l$ is obtained by summing the low-level feature, the upsampled high-level feature, and the reconstructed aligned feature:
\begin{align}
\mathbf{Y}^{l} = \mathbf{X}^{l} + \mathbf{Y}^{l+1}_u + \mathbf{Y}_{\mathrm{recon}}^{l+1}.
\end{align}

\begin{table*}[t]
	\centering
	\caption{Comparisons with state-of-the-art methods on DOTA-v1.0 \cite{dota} dataset with single scale training and testing.}
	\label{tab:dota}
	\setlength{\tabcolsep}{1.0mm}
	\resizebox{\linewidth}{!}{%
		\begin{tabular}{l|c|c|ccccccccccccccc|c}
			\toprule
			Method & Venue&Backbone & PL & BD & BR & GTF & SV & LV & SH & TC & BC & ST & SBF & RA & HA & SP & HC & mAP \\ 
			\midrule
			R. F-RCNN \cite{fasterrcnn} & TPAMI'16 & ResNet50 & 89.40 &81.81 &47.28 &67.44& 73.96 &73.12 &85.03 &90.90& 85.15& 84.90& 56.60& 64.77 &64.70 &70.28 &62.22 &73.17 \\
			R3Det \cite{r3det}&AAAI'21 & ResNet50 & 89.5 & 81.2 & 50.5 & 66.1 & 70.9 & 78.7 & 78.2 & 90.8 & 85.3 & 84.2 & 61.8 & 63.8 & 68.2 & 69.8 & 67.2 & 73.7 \\
			R3Det-GWD \cite{gwd}& ICML'21 & ResNet50 & 88.82 & 82.94 & 55.63 & 72.75 & 78.52 & 83.10 & 87.46 & 90.21 & 86.36 & 85.44 & 64.70& 61.41 & 73.46 & 76.94 & 57.38 & 76.34 \\
			R3Det-KLD \cite{kld}& NIPS'21 & ResNet50 & 88.90 & 84.17 & 55.80 & 69.35 & 78.72 & 84.08 & 87.00 & 89.75 & 84.32 & 85.73 & 64.74 & 61.80 & 76.62 & 78.49 & 70.89 &77.36 \\
			S2ANet \cite{s2anet}&TGRS'21 & ResNet50&89.11& 82.84& 48.37&71.11&78.11&78.39&87.25&90.83&84.90&85.64&60.36&62.60&65.26&69.13&57.94&74.12\\
			RoI Trans. \cite{roitrans}&CVPR'19& ResNet50 & 89.01 & 77.48 & 51.64 & 72.07 & 74.43 & 77.55 & 87.76 & 90.81 & 79.71 & 85.27 & 58.36 & 64.11 & 76.50 & 71.99 & 54.06 &  74.05\\
			ReDet \cite{redet}& CVPR'21 & ReResNet & 88.79 & 82.64 & 53.97 & 74.00 & 78.13 & 84.06 & 88.04 & 90.89 & 87.78 & 85.75 & 61.76 & 60.39 & 75.96 & 68.07 & 63.59 & 76.25 \\
			O-RCNN \cite{orcnn}& IJCV'24 & ResNet50 & 89.46 & 82.12 & 54.78 & 70.86 & 78.93 & 83.00 & 88.20 & 90.90 & 87.50 & 84.68 & 63.97 & 67.69 & 74.94 & 68.84 & 52.28 & 75.87 \\
			ARC \cite{arc}& ICCV'23 & R50-ARC & 89.40 & 82.48 & 55.33 & 73.88 & 79.37 & 84.05 & 88.06 & 90.90 & 86.44 & 84.83 & 63.63 & 70.32 & 74.29 & 71.91 & 65.43 & 77.35 \\
			GRA \cite{gra}& ECCV'24 & R50-GRA & — & — & — & — & — & — & — & — & — & — & — & — & — & — & — & 77.63   \\
			LSKNet \cite{lsknet}& IJCV'24 & LSKNet-S & 89.66 & 85.52 & 57.72 & 75.70 & 74.95 & 78.69 & 88.24 & 90.88 & 86.79 & 86.38 & 66.92 & 63.77 & 77.77 & 74.47 & 64.82 & 77.49 \\
			MutDet \cite{mutdet}& ECCV'24 & ResNet50 & 87.3 & 78.7 & 51.3 & 68.5 & 78.9 & 81.6 & 88.1 & 90.7 & 79.9 & 83.7 & 58.0 & 61.8 & 76.5 & 72.1 &60.8& 74.5 \\
			PKINet \cite{pkinet}& CVPR'24 & PKINet-S & 89.72 & 84.20 & 55.81&77.63&80.25 & 84.45&88.12 & 90.88 & 87.57 & 86.07 & 66.86 & 70.23 & 77.47 & 73.62 & 62.94 & 78.39 \\
			BGHR \cite{bghr}& AAAI'25 & — & 88.21 & 75.65 & 42.83 & 60.17 & 79.39 & 74.99&85.76 & 90.86 & 83.94 & 85.34 & 56.77 & 65.90& 60.18 & 72.45 & 49.11 & 71.44 \\
			ReDiffDet \cite{rediffdet}& CVPR'25 & ResNet50 & 85.45 & 77.84 & 50.39 & 71.63 & 80.90 & 84.85 & 88.77 & 90.88 & 87.00 & 86.08 & 63.30 & 63.43 & 76.01 & 76.03 & 60.16 & 76.18 \\
			ReDiffDet \cite{rediffdet}& CVPR'25 & Swin-T & 82.89 & 82.90 & 52.47 & 70.80 & 80.48 & 85.41 & 88.97 & 90.90 & 86.97 & 87.35 & 60.24 & 65.01 & 76.61 & 74.28 & 64.07 &76.62 \\
			S-RCNN \cite{striprcnn} &AAAI'26 & StripNet-S & 89.56 & 83.86 & 53.65 & 76.27 & 79.31 & 84.93 & 88.24 & 90.89 & 87.69 & 85.88 & 66.15 & 68.11 & 75.73 & 73.32 & 67.79&78.09 \\
			\midrule
			\rowcolor{gray!20}O-RCNN+ours&- & ResNet50 & 89.40&84.15 & 54.66 & 72.23 & 78.42 & 82.89 & 87.86 & 90.89 & 86.26 & 85.24 & 63.95 & 65.58 &75.30& 70.44 & 61.06 & 76.55 \\
			\rowcolor{gray!30}LSKNet+ours& -& LSKNet-S & 89.85 & 85.07 & 55.39 & 77.77 &79.30& 84.72 & 88.56 &90.90&87.74& 86.29 & 68.37&67.74&77.19&74.14&64.30 & 78.49  \\
			\rowcolor{gray!40} S-RCNN+ours&- & StripNet-S & 89.81 & 83.27 & 52.93 & 77.06 & 79.57 & 84.90 & 88.31 & 90.89 & 87.10 & 86.43 & 66.43 & 68.04 & 75.54 & 79.03 &71.48& \textbf{78.72} \\
			\bottomrule
	\end{tabular}}
\end{table*}

\subsection{FAA Head}
\label{sec:faahead}

The FAA Head serves as a new detection head that directly replaces the original head in the Oriented R-CNN \cite{orcnn} framework. 
It takes the RoI-aligned feature $\mathbf{F}_{\text{roi}} $ as input. For each RoI feature, we apply the FAA to align it to a canonical orientation, and specifically, $0^\circ$ in our experiments:
\begin{align}
\mathbf{F}_{\text{inv}} = \mathrm{FAA}(\mathbf{F}_{\text{roi}}; 0^\circ).
\end{align}
The resulting feature $\mathbf{F}_{\text{inv}}$ is rotation-invariant and thus more conducive to classification.

To preserve fine-grained spatial cues that may be lost during alignment, we add the original RoI feature back as a residual:
\begin{align}
\mathbf{F}_{\text{final}} = \mathbf{F}_{\text{inv}} + \mathbf{F}_{\text{roi}}.
\end{align}

We then flatten $\mathbf{F}_{\text{final}}$ into a 1D vector and pass it through two shared fully connected (FC) layers:
\begin{align}
\mathbf{z} = \mathrm{FC} \big( \mathrm{FC}( \mathrm{flatten}(\mathbf{F}_{\text{final}}) ) \big) .
\end{align}

Finally, the resulting feature vector $\mathbf{z}$ is fed into separate classification and regression branches for category prediction and oriented bounding box refinement, respectively.

\section{Experiments}
\label{sec:experiments}

In this section, we first introduce our experimental settings, including the datasets we used and how we implement our experiments. Then we will exhibit our results on these datasets. In the end of this section is our analysis, consisting of ablation studies, the effectiveness of our FAA Head compared with the latest methods and the analysis of high-IoU performance, which is \textit{de facto} a realistic question.

\subsection{Experimental Settings}
\noindent\textbf{Datasets.}
We conduct our experiments on three popular remote sensing object detection datasets: DOTA-v1.0 \cite{dota}, DOTA-v1.5 \cite{dota} and HRSC2016 \cite{hrsc2016}. 

\begin{itemize}
    \item \textbf{DOTA-v1.0} \cite{dota} is a large-scale datasets for remote sensing object detection, which includes 2,806 images, 188,282 instances and 15 categories. The dataset is comprised of 1,411, 458, and 937 images for \texttt{train}, \texttt{val} and \texttt{test} sets, respectively.
    \item \textbf{DOTA-v1.5} \cite{dota} is an extended version of DOTA-v1.0, which is released for DOAI Challenge 2019. It contains the same 2,806 images but with 403,318 instances and 16 categories, including the refined annotations for vehicle instances and a new category named \texttt{Container\ Crane} (CC). Moreover, this iteration has a substantial increase of the minuscule instances that are less than 10 pixels, which is more challenging.
    \item \textbf{HRSC2016} \cite{hrsc2016} is a dataset for ship detection in remote sensing imagery, consisting of 1,061 images and 2,976 ship instances annotated by a horizontal bounding box and an orientation angle.  The images are split into 436, 181 and 444 for \texttt{train}, \texttt{val} and \texttt{test} sets.

\end{itemize}

\noindent\textbf{Evaluation Metrics.}
Followed by previous works \cite{orcnn,lsknet,pkinet,striprcnn}, we employ the mean Average Precision (mAP) as our evaluation metrics for the two DOTA \cite{dota} datasets. As for HRSC2016 \cite{hrsc2016}, we evaluate our results with PASCAL VOC2007 \cite{voc07}/VOC2012 \cite{voc12} metrics with AP50, AP75 and mAP reported, according to the preceding work \cite{gra}.

\begin{table*}[ht]
	\centering
	\caption{Comparisons with state-of-the-art methods on DOTA-v1.5 \cite{dota} dataset with single scale training and testing.}
	\label{tab:dota15}
	\setlength{\tabcolsep}{1.0mm}
	\resizebox{\linewidth}{!}{%
		\begin{tabular}{l|c|cccccccccccccccc|c}
			\toprule
			Method &Backbone & PL & BD & BR & GTF & SV & LV & SH & TC & BC & ST & SBF & RA & HA & SP & HC & CC & mAP \\ 
			\midrule
			RetinaNet-O \cite{retinanet} & ResNet50 & 71.43 &77.64& 42.12 &64.65& 44.53 &56.79 &73.31 &90.84 &76.02 &59.96 &46.95& 69.24& 59.65 &64.52 &48.06 &0.83 &59.16 \\
			R. F-RCNN \cite{fasterrcnn} & ResNet50 & 71.89 &74.47& 44.45 &59.87 &51.28 &68.98 &79.37 &90.78& 77.38& 67.50 &47.75 &69.72 &61.22 &65.28 &60.47 &1.54& 62.00 \\
			Mask R-CNN \cite{maskrcnn} & ResNet50 & 76.84& 73.51& 49.90& 57.80& 51.31& 71.34& 79.75& 90.46& 74.21& 66.07 &46.21& 70.61 &63.07& 64.46 &57.81& 9.42 &62.67 \\
			HTC \cite{htc} & - & 77.80 &73.67& 51.40& 63.99& 51.54& 73.31& 80.31& 90.48 &75.12 &67.34 &48.51 &70.63 &64.84 &64.48 &55.87 &5.15& 63.40 \\
			RoI Trans. \cite{roitrans} & ResNet50 & 71.70& 82.70& 53.00 &71.50& 51.30 &74.60 &80.60 &90.40 &78.00 &68.30 &53.10 &73.40 &73.90 &65.60& 56.90 &3.00 & 65.50\\
			O-RCNN \cite{orcnn} & ResNet50 & 79.68&81.72 & 53.36 & 72.17 & 52.26 & 76.47 & 81.00 & 90.86 & 78.93 & 68.18 & 65.05 & 72.17 &67.68& 65.23 & 56.11 & 7.42 & 66.77 \\
			ReDet \cite{redet} & ReResNet & 79.20 &82.81& 51.92 &71.41 &52.38 &75.73 &80.92 &90.83 &75.81 &68.64 &49.29 &72.03 &73.36 &70.55 &63.33 &11.53 &66.86 \\
			S-RCNN \cite{striprcnn}  & StripNet-S & 80.31 & 82.24 & 53.91 & 76.61 & 52.39 & 81.55 & 87.82 & 90.89 & 82.44 & 66.32 & 61.58 & 73.27 & 75.04 & 65.84 & 73.42&13.85&69.84 \\
			LSKNet  \cite{lsknet} & LSKNet-S &  72.05 &84.94& 55.41& 74.93& 52.42& 77.45& 81.17 &90.85 &79.44 &69.00 &62.10 &73.72& 77.49 &75.29 &55.81 &42.19 &70.26 \\
			PKINet \cite{pkinet} & PKINet-S & 80.31 &85.00& 55.61& 74.38 &52.41& 76.85 &88.38 &90.87 &79.04 &68.78& 67.47& 72.45 &76.24 &74.53 &64.07 &37.13 &71.47 \\
			\midrule
			\rowcolor{gray!20} O-RCNN+ours & ResNet50 & 79.62&81.14 & 53.96 & 73.78 & 52.20 & 76.80 & 81.17 & 90.90 & 78.47 & 68.55 & 64.21 & 73.72 &68.26& 64.64 & 54.09 & 12.85 & 67.14 \\
			\rowcolor{gray!30} S-RCNN+ours & StripNet-S & 80.14 & 82.02 & 51.89 & 76.42 & 52.52 & 81.45 & 88.21 & 90.88 & 85.76 & 68.52 & 62.34 & 72.68 & 74.01 & 74.97 &71.20& 32.10 & 71.57 \\
			\rowcolor{gray!40} LSKNet+ours& LSKNet-S & 80.69 & 85.34 & 55.42 & 77.91 &52.64& 82.26 & 88.48 & 90.85 & 86.38 & 69.32 & 62.04 & 74.11 & 75.83 & 74.70 & 73.18 & 27.30 & \textbf{72.28}  \\
			\bottomrule
	\end{tabular}}
\end{table*}

\noindent\textbf{Implementation details.}
    All the experiments are implemented under the  MMRotate \cite{zhou2022mmrotate} toolbox developed by Openmmlab. Followed by  previous works \cite{orcnn,lsknet,pkinet,striprcnn}, for DOTA-v1.0 \cite{dota} and DOTA-v1.5 \cite{dota}, the input size is set to $1024 \times 1024$, and for HRSC2016, the input size is set to $800 \times 800$. For DOTA-v1.0 \cite{dota} and DOTA-v1.5 \cite{dota}, training is conducted for 16 epochs on \texttt{trainval} set. Models from epochs 12 to 16 are evaluated on \texttt{test} set, and the best-performing one is selected as the final result. For HRSC2016 \cite{hrsc2016}, training is conducted for 36 epochs and the last weight is used for testing. During the training progress, we use AdamW \cite{adamw} as the optimizer with the weight decay set to 0.05. The initial learning rate is set to 0.0004 for HRSC2016 \cite{hrsc2016} and 0.0001 for the two DOTA \cite{dota} datasets. With a batch size of 2, all training and testing are performed on a single RTX 3090 GPU.

\subsection{Main Results}

We first add our FAAFusion and FAA Head to three representative state-of-the-art methods within the Oriented R-CNN \cite{orcnn} framework on the datasets mentioned above. Among these methods, ResNet50 \cite{resnet} is the most classic and popular backbone, which is widely used or modified not only in object detection tasks but also in other computer vision tasks. LSKNet-S \cite{lsknet} is recently the most powerful in both detection and segmentation, and StripNet-S \cite{striprcnn} is the latest state-of-the-art method.

\noindent\textbf{Results on DOTA-v1.0 \cite{dota}.}
 As shown in Table \ref{tab:dota}, we choose ResNet50 \cite{resnet}, LSKNet-S \cite{lsknet} and StripNet-S \cite{striprcnn} as the backbone, and add our FAAFusion and FAA Head into the model. Specially, for Strip R-CNN \cite{striprcnn} who has designed the Strip Head, we replace their Strip Head with our FAA Head. Based on these models, our methods achieve new improvements of +0.68\%, +1.00\% and +0.63\%, respectively. Compared to current state-of-the-art method PKINet \cite{pkinet}, our results using StripNet-S \cite{striprcnn} achieve new sota result of 78.72\%, with 0.33\% higher on mAP.

\noindent\textbf{Results on DOTA-v1.5 \cite{dota}.}
 Similarly, we choose ResNet50 \cite{resnet}, LSKNet-S \cite{lsknet} and StripNet-S \cite{striprcnn} as the backbone, and add our FAAFusion and FAA Head into the model. For Strip R-CNN \cite{striprcnn}, we replace their Strip Head with our FAA Head as before. As shown in Table \ref{tab:dota15}, based on these models, our methods achieve new improvements of 67.14\%, 71.57\% and 72.28\%, with the increase of 0.37\%, 1.73\% and 2.02\% on each models, respectively. Compared to current two-stage sota detector PKINet \cite{pkinet}, our results using LSKNet-S \cite{striprcnn} achieve new sota result of 72.28\%, with 0.81\% higher on mAP. 

\noindent\textbf{Results on HRSC2016 \cite{hrsc2016}.}
As shown in Table \ref{tab:hrsc}, we integrate our FAAFusion and FAA Head into three representative oriented object detectors: Oriented R-CNN \cite{orcnn}, LSKNet \cite{lsknet} and Strip R-CNN \cite{striprcnn}. Followed by these works, we report the AP50 within the evaluation protocols of PASCAL VOC2007 \cite{voc07} and VOC2012 \cite{voc12}. Moreover, to compare the detection performance of the model under stricter criteria, we also report the AP75 as well as mAP results of COCO \cite{coco} style. Under the VOC2007 \cite{voc07} metric, our method consistently improves performance across all baselines, achieving mAP gains of 2.17\%, 1.81\% and 1.23\%, respectively. Under the VOC2012 \cite{voc12} metric, our approach boosts mAP by 1.72\% for Oriented R-CNN \cite{orcnn}, 2.34\% for LSKNet \cite{lsknet}, and maintains competitive performance of 0.43\% higher for Strip R-CNN \cite{striprcnn}. 

\begin{table}[t]
\centering
\caption{Comparisons with three representative state-of-the-art methods on HRSC2016 \cite{hrsc2016} dataset. The values of APs are calculated with PASCAL VOC2007 \cite{voc07}/VOC2012 \cite{voc12} metrics and the mAPs are reported with COCO \cite{coco} styles. All the models are retrained and tested on a single RTX 3090.  The Params and FLOPs are calculated with MMRotate \cite{zhou2022mmrotate} toolbox.}
\label{tab:hrsc}
\resizebox{\linewidth}{!}{%
\begin{tabular}{l|c|c|c|c|c}
\toprule
Method&\#Params$\downarrow$ &\#FLOPs$\downarrow$&AP50$\uparrow$&AP75$\uparrow$&mAP$\uparrow$\\
\midrule
\multicolumn{6}{l}{PASCAL VOC2007 \cite{voc07}} \\
\midrule
O-RCNN \cite{orcnn}& 41.13M & 134.46G &89.7&79.5&64.77\\
\rowcolor{gray!20}O-RCNN+ours& 63.27M & 140.70G &89.8&80.0&\textbf{66.94}\\
\midrule
LSKNet \cite{lsknet}& 30.96M & 111.42G &90.2&87.9&68.78\\
\rowcolor{gray!20}LSKNet+ours& 48.34M & 114.89G &90.6&89.8&\textbf{70.74}\\
\midrule
S-RCNN \cite{striprcnn}& 45.12M & 157.19G &89.5&78.8&69.18\\
\rowcolor{gray!20}S-RCNN+ours& 49.05M & 115.91G & 90.0& 78.6& \textbf{70.41}\\
\midrule
\midrule
\multicolumn{6}{l}{PASCAL VOC2012 \cite{voc12}} \\
\midrule
O-RCNN \cite{orcnn}& 41.13M & 134.46G &96.3&84.1&67.80\\
\rowcolor{gray!20}O-RCNN+ours& 63.27M & 140.70G &95.9&85.9&\textbf{69.52}\\
\midrule
LSKNet \cite{lsknet}& 30.96M & 111.42G &97.8&89.2&72.37\\
\rowcolor{gray!20}LSKNet+ours& 48.34M & 114.89G &98.3&91.8&\textbf{74.71}\\
\midrule
S-RCNN \cite{striprcnn}& 45.12M & 157.19G  &97.7&87.6&72.56\\
\rowcolor{gray!20}S-RCNN+ours& 49.05M & 115.91G & 98.2& 89.7& \textbf{72.99}\\
\bottomrule
\end{tabular}}
\end{table}

\subsection{Analysis}

\noindent\textbf{Ablation study of the two proposed modules.}
To measure the effectiveness of FAAFusion and FAA Head, we conduct ablation experiments on Oriented R-CNN \cite{orcnn} framework with LSKNet-S \cite{lsknet} as the backbone. We replace the element-wise add with our FAAFusion when fusing the third-level and second-level features in FPN \cite{fpn} and change the original detection head to FAA Head. The FAA Head follows the detection head design of Oriented R-CNN \cite{orcnn} framework, using two shared fully connected layers before classification and regression. But different from the original output dimension of the first fully connected layer that is set to $1024$, the output dimension of our FAA Head is set to $1024 + 256$. As shown in Table \ref{tab:ablation}, we evaluate the impact of the FAAFusion and FAA Head on DOTA-v1.0 \cite{dota} dataset. Results show that FAAFusion and FAA Head respectively improve the performance of LSKNet \cite{lsknet} by 0.42\% and 0.78\%, which means both of our two modules could make contributions to the results.

\begin{table}[t]
\centering
\caption{Ablation study. We conducted the experiment with LSKNet-S \cite{lsknet} as the backbone. The Params and FLOPs are calculated with MMRotate \cite{zhou2022mmrotate} toolbox.}
\label{tab:ablation}
\resizebox{\linewidth}{!}{%
\begin{tabular}{c|c|c|c|c}
\toprule
FAAFusion&FAA Head&\#Params$\downarrow$ &\#FLOPs$\downarrow$& mAP$\uparrow$\\
\midrule
\ding{56} & \ding{56} & 30.98M & 173.68G & 77.49\\
\ding{56} & \ding{52} & 48.35M & 177.15G & 78.27\\
\ding{52} & \ding{56} & 32.18M & 175.59G & 77.91\\
\midrule
\ding{52} & \ding{52} & 49.56M & 179.06G & \textbf{78.49}\\
\bottomrule
\end{tabular}}
\end{table}

\noindent\textbf{Comparison with the latest work of detection head.}
To evaluate the effectiveness of our FAA Head, we verify the detection head on three representative backbones: ResNet50 \cite{resnet}, LSKNet-S \cite{lsknet} and StripNet-S \cite{striprcnn}. Specifically, we choose the latest detection head, the Strip Head \cite{striprcnn}, together with the original detection head, as our comparison. As shown in Table \ref{tab:headanalysis}, across all these three detectors, the FAA Head achieves the highest mAP value. Compared with the Strip Head \cite{striprcnn}, FAA Head achieves higher mAP with very close parameters but much lower computational cost. These results indicate that the FAA Head can consistently improve detection performance without substantially increasing the computational cost.

\begin{table}[t]
\centering
\caption{Comparison with other detecting head on other remote sensing object detectors. The Params and FLOPs are calculated with MMRotate \cite{zhou2022mmrotate} toolbox.}
\label{tab:headanalysis}
\resizebox{\linewidth}{!}{
\begin{tabular}{c|c|c|c|c}
\toprule
Method&Detecting Head&\#Params$\downarrow$ &\#FLOPs$\downarrow$& mAP$\uparrow$\\
\midrule
\multirow{3}*{O-RCNN \cite{orcnn}}  & Original Head & 41.14M & 211.43G & 75.81\\
~ & Strip Head & 55.82M & 258.35G & 76.11\\
~ & FAA Head & 58.51M & 214.90G & \textbf{76.18}\\
\midrule
\midrule
\multirow{3}*{LSKNet \cite{lsknet}}  & Original Head & 30.98M & 173.68G & 77.49\\
~ & Strip Head & 45.65M & 220.60G & 78.04\\
~ & FAA Head & 48.35M & 177.15G & \textbf{78.27}\\
\midrule
\midrule
\multirow{3}*{S-RCNN \cite{striprcnn}}  & Original Head & 30.46M & 171.79G & 77.03\\
~ & Strip Head & 45.14M & 218.71G & 78.09\\
~ & FAA Head & 47.83M & 175.26G & \textbf{78.52}\\
\bottomrule
\end{tabular}}
\end{table}

\begin{figure}[t]
\centering
\includegraphics[width=0.49\linewidth]{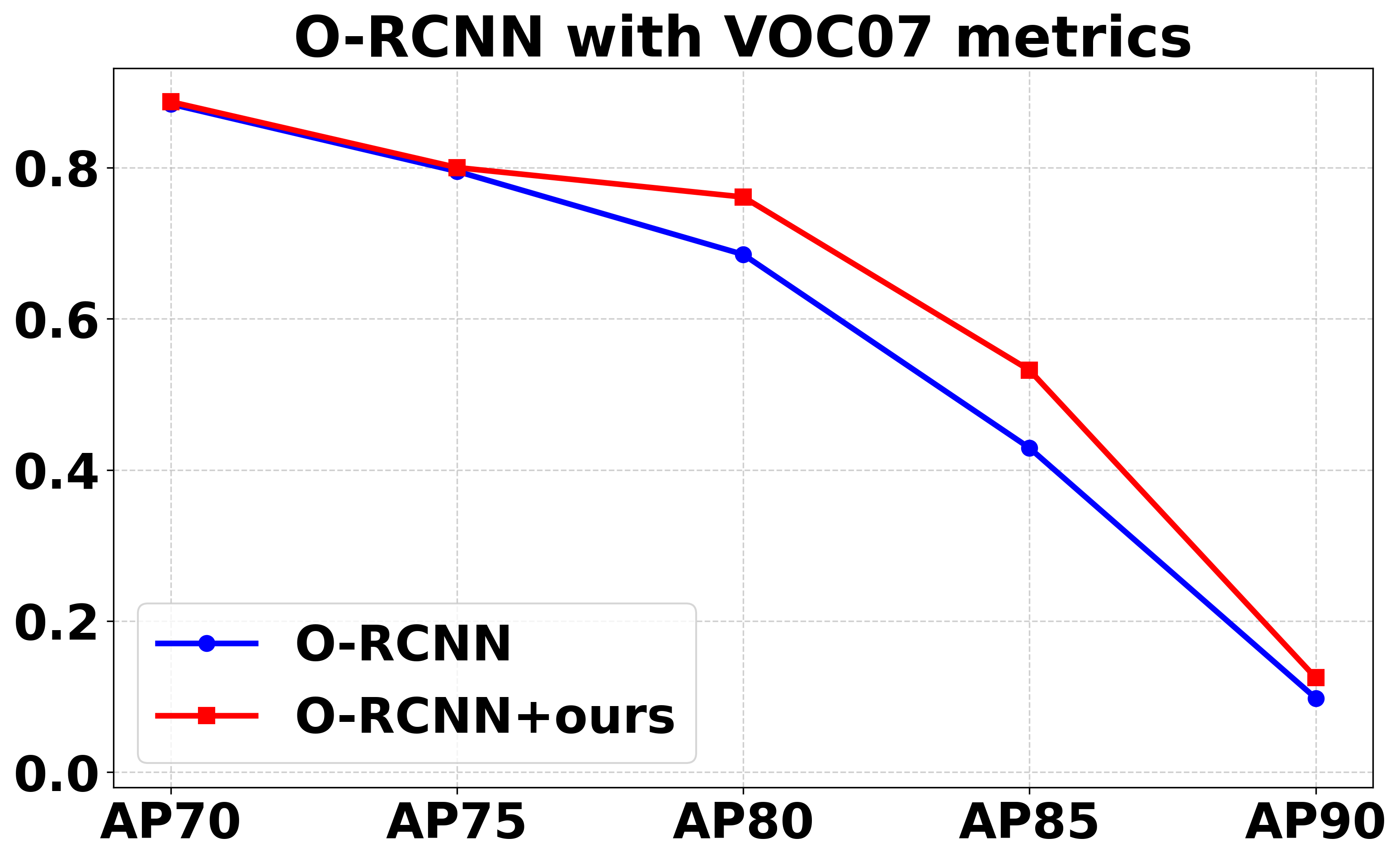}\hfill
\includegraphics[width=0.49\linewidth]{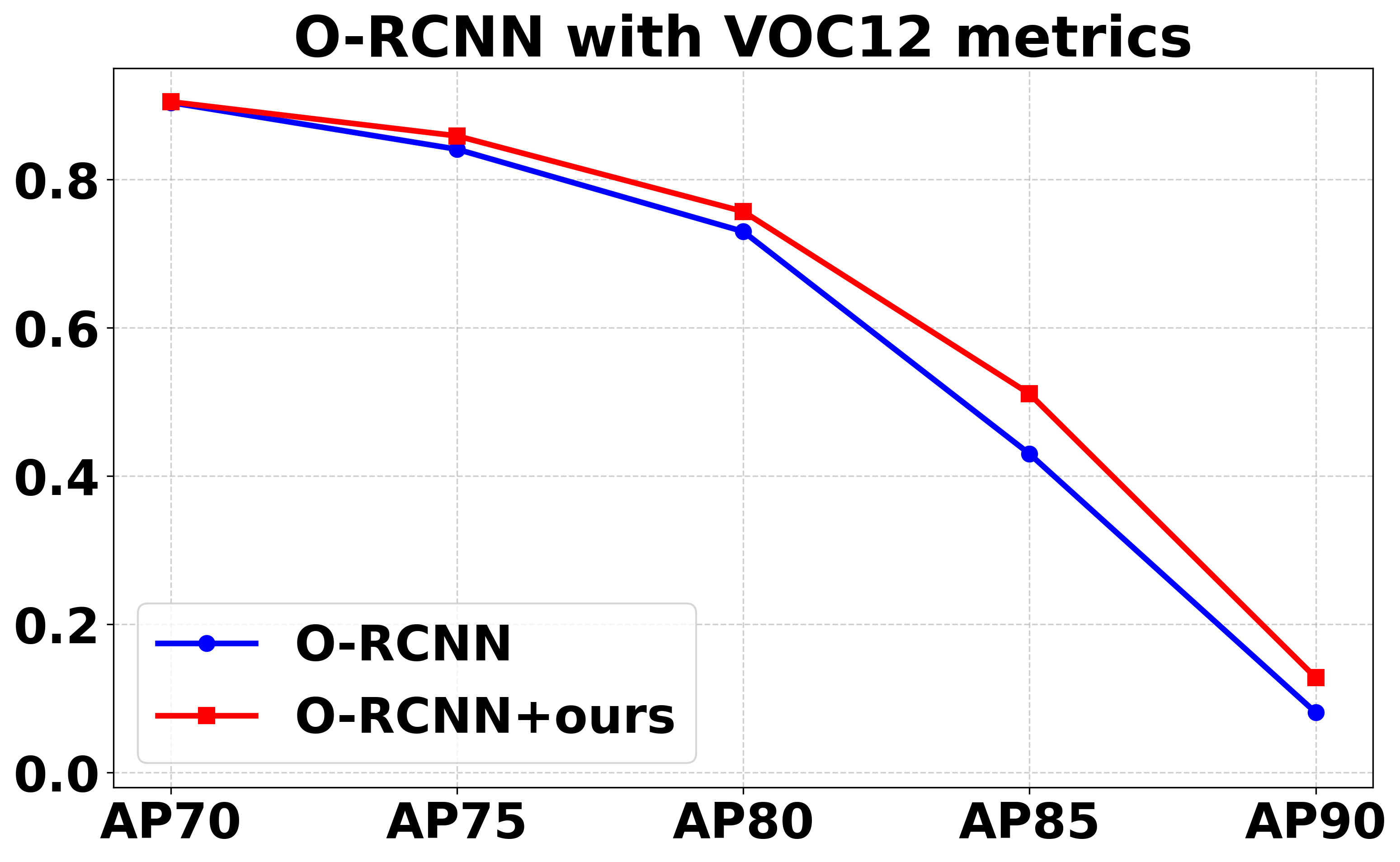}\\

\includegraphics[width=0.49\linewidth]{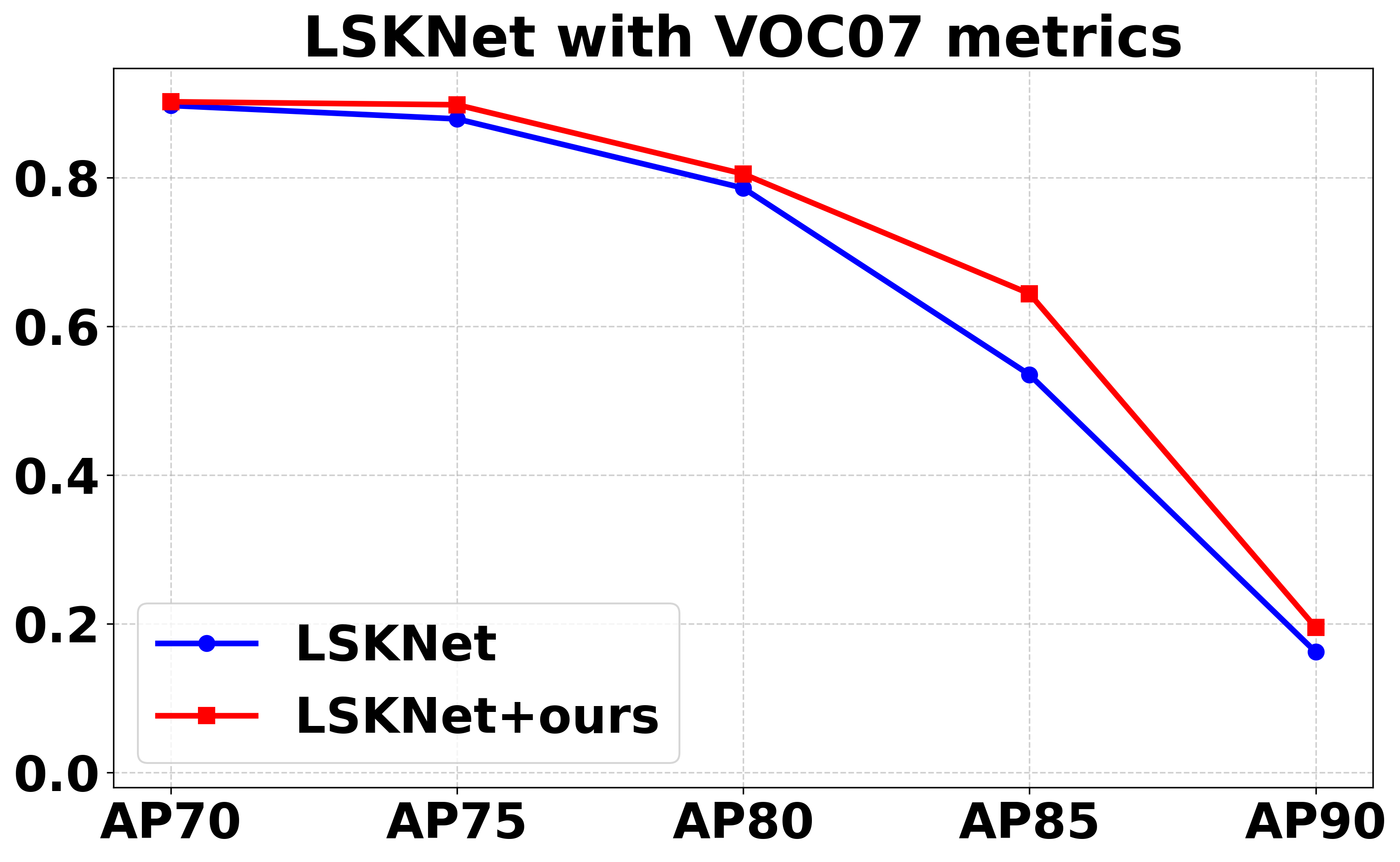}\hfill
\includegraphics[width=0.49\linewidth]{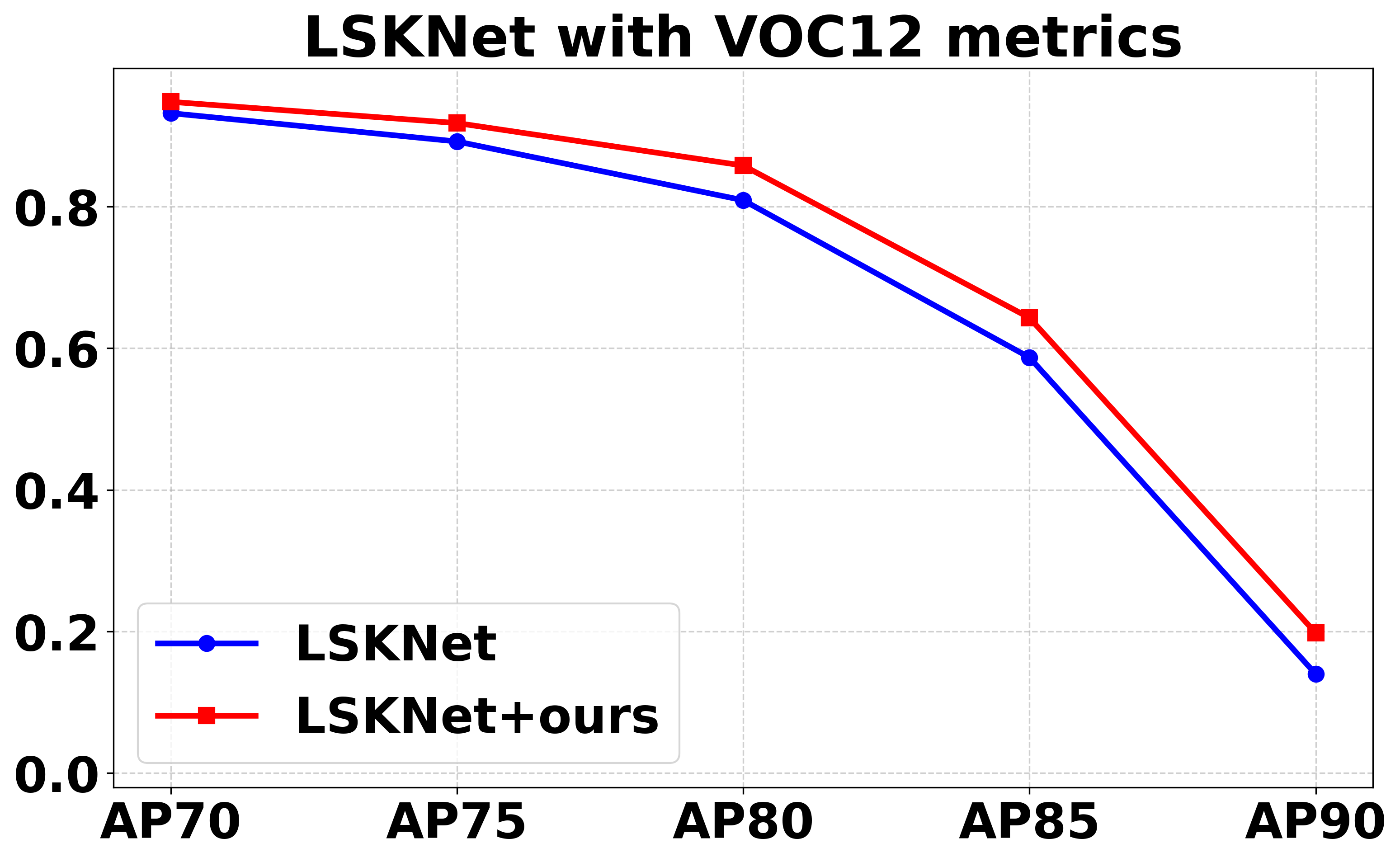}

\caption{Effectiveness of our method at Oriented R-CNN \cite{orcnn} and LSKNet \cite{lsknet} compared to the original methods on HRSC2016 \cite{hrsc2016} dataset. In the high IoU threshold range, our method performs better. The AP values are calculated with PASCAL VOC2007 \cite{voc07} and VOC2012 \cite{voc12} metrics.}
\label{fig:4cmps}
\end{figure}

\noindent\textbf{Analysis of high-IoU performance.}
In the current evaluation protocol for rotated ship detection, the results of AP50 are almost saturated. Differences among methods are tiny and hide true fine-localization ability. Once the IoU threshold rises above 0.7, the performance of mainstream detectors drops sharply, showing that their geometric modeling is still coarse. Tasks such as ship detection, however, highly demand strict box fitting and accurate orientation, because small misalignments could lead to later recognition or ranging errors. Testing robustness in the high-IoU range therefore removes the artificial advantage given by loose metrics and directly checks whether a model is well-performed in precision-critical scenes. Driven by this consideration, we compute AP of high IoU thresholds from 0.70 to 0.90 and plot the curves. As illustrated in Figure \ref{fig:4cmps}, the advantage of our method becomes increasingly evident at higher IoU thresholds, demonstrating its superior capability in precise orientation modeling.

\section{Conclusion}
\label{sec:conclusion}

In this work, we propose Fourier Angle Alignment (FAA), leveraging fourier rotation equivariance and spectral alignment of rectangles to better model orientation cues in oriented object detection tasks. We theoretically show that the spectral energy of a rectangular object concentrates perpendicular to its major axis, and use this insight to estimate dominant orientations in the frequency domain and align features accordingly. We integrate FAA into both neck and the detecting head, designing FAAFusion and FAA Head, which enhance orientation consistency across scales and improve rotation invariance of RoI features with minimal computational overhead. Experiments on three benchmark datasets demonstrate the efficacy of our approach in remote sensing object detection. In the future, it is worth exploring the generalization of FAA to other vision tasks such as instance segmentation or remote sensing changing detection, and investigate more efficient frequency-domain alignment strategies for real-time applications.

 \section*{Acknowledgement}

This work is supported by the National Natural Science Foundation of China (62331006), and the Fundamental Research Funds for the Central Universities.
 {
     \small
     \bibliographystyle{ieeenat_fullname}
     \bibliography{main}
 }



\end{document}